\documentclass[10pt]{article}
\usepackage{fullpage}
\usepackage[tight,scriptsize]{subfigure}
\def\cut#1{{}}
\def\cancut#1{{#1}}
\def\h{{\kappa}}
\def\eg{{\em e.g.,}}
\def\ie{{\em i.e.,}}
\def\etc{{\em etc.}}
\usepackage{times}
\usepackage{epsfig}
\usepackage{graphicx}
\usepackage{amsmath}
\usepackage{amssymb}
\usepackage{amsthm}
\usepackage[pagebackref=true,breaklinks=true,letterpaper=true,colorlinks,bookmarks=false]{hyperref}
\def\real{\mathbb{R}}

\newcommand{\be}{\begin{equation}}
\newcommand{\ee}{\end{equation}}
\newcommand{\ba}{\left[ \begin{array}}
\newcommand{\ea}{\end{array} \right]}
\newcommand{\bea}{\begin{eqnarray}}
\newcommand{\eea}{\end{eqnarray}}

\def\real{\mathbb{R}}

\newtheorem{claim}{Claim}

\begin{document}
\title{On the Design and Analysis of Multiple View Descriptors} 
\author{Jingming Dong \and Jonathan Balzer \and Damek Davis \and Joshua Hernandez \and Stefano Soatto\thanks{We wish to thank Virginia Estellers for discussions and suggestions. Research supported by ARO, ONR and AFOSR.}}
\maketitle
\begin{abstract}
We propose an extension of popular descriptors based on gradient orientation histograms (HOG, computed in a single image) to multiple views. It hinges on interpreting HOG as a conditional density in the space of sampled images, where the effects of nuisance factors such as viewpoint and illumination are marginalized. However, such marginalization is performed with respect to a very coarse approximation of the underlying distribution. Our extension leverages on the fact that multiple views of the same scene allow separating intrinsic from nuisance variability, and thus afford better marginalization of the latter. The result is a descriptor that has the same complexity of single-view HOG, and can be compared in the same manner, but exploits multiple views to better trade off insensitivity to nuisance variability with specificity to intrinsic variability. We also introduce a novel multi-view wide-baseline matching dataset, consisting of a mixture of real and synthetic objects with ground truthed camera motion and dense three-dimensional geometry. 
\end{abstract}
\section{Introduction}

Images of a particular object or scene depend on its {\em intrinsic properties} (\eg shape, reflectance) and on {\em nuisance factors} (\eg viewpoint, illumination, partial occlusion, sensor characteristics, \etc) that have no bearing on its intrinsic properties but nevertheless affect the data. A (feature) {\em descriptor} is a function of the data designed to {remove or reduce nuisance variability while preserving intrinsic variability.} 
The {\em ideal descriptor} would be the probability density of the data (images), conditioned on the intrinsic factors, with nuisance factors integrated out (marginalized).\footnote{That would be the likelihood function, which is a minimal sufficient statistic.} This would enable evaluating the likelihood that any image is generated by a given object or scene. This is key to many visual decisions (detection, recognition, localization, categorization) that hinge on evaluating the probability that (local regions) in two images {\em correspond}, \ie are generated by the same portion of the underlying scene.

\subsection{Contributions}
The first observation we offer is that an extension of commonly used descriptors such as SIFT/HOG \cite{lowe04distinctive,dalalT05}\footnote{And their many variants, which we refer to collectively as HOG.} (\ref{eq-h}), {\em can be interpreted as such a class-conditional density, but with nuisance factors marginalized with respect to a coarse approximation} of the underlying nuisance distribution (Claim \ref{claim-bad}). Such an approximation is necessarily poor, as {\em a single image does not afford the ability to separate intrinsic from nuisance variability} (Sec. \ref{sect-notation}). Thus, {\em any} descriptor constructed from a single view fails to be shape-discriminative (Claim \ref{claim-no-shape}).

The second observation is that {\em multiple views of the same underlying scene (\eg a video) allow us to attribute the variability in the data to nuisance factors,} if the underlying scene is static at the time-scale of capture. Thus, our {\em working hypothesis} is that one ought to be able to leverage multiple views to construct {\em better descriptors}, \ie better approximations the class-conditional density. 

Our {\em main contribution} is to show how this can be done in two ways: Using a sample approximation of the nuisance distribution, obtained via a {\em tracker}, leading to Multi-view HOG (MV-HOG, Sec. \ref{sect-mv-hog}), or using a point estimate of the intrinsic factors, obtained via multi-view stereo or other {\em reconstruction} method, leading to Reconstructive HOG (R-HOG, Sec. \ref{sect-marginalization}). The two coincide under {\em sufficient excitation} conditions \cite{bitmead84} on the samples. In either case, the result is a descriptor that has {\em the same run-time and storage complexity of standard (single-view) HOG}, but better separates intrinsic from nuisance variability. When only one {\em ``training image''} is given, both reduce to single-view HOG. 

Comparison with a (single, or multiple) {\em ``test image''} can be performed in three ways, corresponding to different interpretations of the descriptors (Sec. \ref{sect-comparison}): As {\em expectations} (statistics), using a distance in the embedding (linear) space, as customary; as {\em distributions}, using any distance of divergence between probability densities; as {\em likelihoods}, by testing the likelihood of a (test) image under the probability model of the underlying scene, encoded in the training images. 

To {\em empirically validate} the descriptors proposed, we need a dataset that provides multiple training images (\eg video), and a plurality of test images obtained under different viewpoint, illumination, partial occlusion \etc. To make comparison of {\em descriptors} independent of the detection mechanism, we need {\em ground truth shape}, to establish correspondence between training and test images. Despite our best efforts, we have not been able to identify such a dataset among those commonly used for benchmarking descriptors. Therefore, another contribution we offer is a {\em new dataset for multi-view wide-baseline matching} (Sec. \ref{sect-dataset}), comprising a variety of real and synthetic objects, with ground truth camera motion and object shape. Both the dataset and our code will be publicly released upon completion of the anonymous review process. Strengths and weaknesses of our method are discussed in Sec. \ref{sect-discussion}.

\subsection{Related Work}

The literature offers a multitude of local single-view descriptors and empirical tests, \eg \cite{mikolajczyk04comparison} as of 2004. More recent examples include \cite{bay2006surf,CHOG,BRIEF,rublee2011orb,DAISY,BRISK,FREAK}; \cite{brown2011discriminative,winder2007learning} couple learning with optimization to minimize classification error;  \cite{chatfield2011devil,vedaldiS08ICVSS} analyze important implementation details. There are relatively few examples of local multi-view descriptors: \eg \cite{delponte2007importance} combines spatial (averaged SIFT) and temporal information (co-visibility), \cite{grabner2005object} learns descriptors by feature selection from trajectories of key points. \cite{meltzerGS04} uses kernel principal component analysis to learn variations among tracked patches for wide baseline matching.\cite{leeS10} learns the best template by taking the modes of learned distributions over time.

{Technically, a descriptor is a statistic (a deterministic function of the data), designed to be insensitive to nuisance variability and yet retain intrinsic variability. The process of finding the optimal tradeoff between two properties that can be described probabilistically in a conditional model that has been formalized by the Information Bottleneck (IB) \cite{tishbyPB00}, a generalization of the notion of sufficient statistics. Thus, ideally, we would like to find the sufficient statistics of images for the purpose of classification under changes of viewpoint, illumination and partial occlusions. Unfortunately, this problem is intractable.} A recent trend of learning descriptors {\em ab-ovo} \cite{ranzato2007unsupervised,memisevic2010learning} is interesting but the results hard to analyze.\cut{ approximate IB, but this link has never been established. Furthermore, recent results \cite{sundaramoorthiPVS09} point to the impossibility of ``learning away'' nuisance variability due to viewpoint and illumination, a fact that is true {\em a fortiori} in the presence of occlusions.} At the opposite end of the modeling spectrum, statistical approaches prevalent in the nineties \cite{grenanderM94} have failed to translate into state-of-the-art methods. Some approaches \cite{lecun04learning,mallatB11,poggio11, von2012d} are positioned in between, with some stages engineered and the rest learned. Nevertheless, some of the most fundamental questions on how to relate these approaches are still largely open.

\section{Methodology}

\subsection{Notation and assumptions}
\label{sect-notation}

Images $\{I_t\}_{t=1}^T$ of a static object (or ``scene''), with $I_t: D\subset \real^2 \rightarrow \real^+; \ x \mapsto I_t(x)$ and $\nabla I_t : D \rightarrow \real^2; \ x \mapsto \nabla I_t(x)$ its spatial gradient, obtained under different viewpoints and illuminations, can be represented as domain deformations (``warpings'') $w_t:\real^2\rightarrow \real^2$ and range deformations (``contrast'') $\h_t: \real\rightarrow \real$ of a common radiance function (informally ``texture map'' or ``albedo'') $\rho: S\rightarrow \real^+$ defined on the surface (``shape'') $S\subset \real^3$. With an abuse of notation, we can define $\rho$ on $D \subset \real^2$ by back-projecting it onto the surface: $\rho(\pi^{-1}_S(x)) \doteq \rho(x)$, where $\pi_S^{-1}(x)$ is the point of first intersection of the pre-image of $x$ under perspective projection $\pi:\real^3 \rightarrow \real^2; X \mapsto \pi(X) = X/X_3$ (the line through the optical center and the pixel with coordinates $x \in D \subset \real^2$) with the surface $S$. If the vantage point is represented by a Euclidean reference frame $g_t = (R_t, T_t) \in SE(3)$ with position $T_t\in \real^3$ and orientation $R_t\in SO(3)$ relative to some global frame, then the warping
\be
w_t(x) = \pi \circ g_t \circ \pi^{-1}_S(x)
\label{eq-w}
\ee
entangles the shape $S$ (an intrinsic property of the scene) and the viewpoint $g_t$ (a nuisance factor in the data formation process). The composition of function is defined as $f\circ g(x) \doteq f(g(x))$. Then the image is generated via
\be
I_t(x) = \h_t \circ \rho \circ w_t (x) + n_t(x)
\label{eq-I}
\ee
up to a residual $n$ (``noise'') assumed white, zero-mean, homoscedastic and Gaussian, with a variance $\epsilon > 0$: $n_t(x) \stackrel{\rm IID}{\sim} {\cal N}_\epsilon(n)$, where ${\cal N}_b(a) \doteq \frac{1}{\sqrt{2\pi}b}\exp(-\frac{\| a \|^2}{\sigma^2})$ and IID stands for independent and identically distributed.

This data formation model is only valid for Lambertian scene under constant diffuse illumination, away from occluding boundaries. We will therefore assume these conditions henceforth.\cut{ Note that the joint distribution of pixel values in the image $I(x)$ can be rather complicated, but the {\em conditional} density $p(I(x) | \h_t, \rho, w_t)$ is spatially IID.}
The {\em ideal descriptor} would be the probability density of any image $I$, {\em conditioned on the intrinsic factors} $\rho, S$, and marginalized with respect to the nuisances $\h_t, g_t$: 
\be
p(I|\rho, S) = \int {\cal N}_\epsilon(I - \h_t \circ \rho \circ w_t)dP(\h_t,w_t | \rho, S).
\label{eq-marg}
\ee
This is the {\em likelihood function}, which is a minimal sufficient statistic. Although the distribution of an image is extremely complex, {\em conditioned} on the scene and nuisances it is IID under the assumptions of the model (\ref{eq-I}).

\subsection{HOG as a conditional density} 

This section derives a continuous gradient orientation distribution, from which various descriptors such as HOG and SIFT are obtained by sampling (App. \ref{sect-sparse-hog}).\cancut{ This immediately suggests a point-wise inversion formula, in the sense of maximum likelihood.} Let $M: D \rightarrow \real^+; \ x \mapsto M(x)$ be the gradient magnitude,
\[
M(x) \doteq \| \nabla I(x) \|
\]
and $G: D \backslash \{x \ | \ M(x) = 0 \} \rightarrow {\mathbb S}^1; \ x \mapsto G(x)$ the gradient orientation, represented by a unit-norm vector
\[
G(x) \doteq \frac{\nabla I(x)}{ \| \nabla I(x) \|}
\]
$\angle G$ is the angle formed with the abscissa. The (un-normalized) density of gradient orientations (DOG) is a function $h: D \times {\mathbb S}^1 \rightarrow \real^+$ with parameters $0 < \epsilon < \pi$ and $0 < \sigma < \sqrt{|D|}$ representing the angular and spatial kernel widths: 
\be
\boxed{h(x, \theta) = \int_{\real^2} {\cal N}_\epsilon\left( \theta - \angle G(y) \right) {\cal N}_\sigma \left( x - y \right) M(y) dy} ~~~ {\rm HOG}
\label{eq-h}
\ee
where ${\cal N}_\epsilon(\theta)$ is an angular Gaussian\cancut{\cite{watson1983statistics}}. Popular descriptors, such as \cite{dalalT05,lowe04distinctive}\footnote{{\em E.g.,} \cite{lowe04distinctive} quantizes orientation $\theta$ into $8$ bins and samples $x$ on a $4\times 4$ lattice, using a bi-linear kernel instead of a Gaussian.}, are sampled versions of (\ref{eq-h}) with different kernels. The measure 
\[\mu(dy) \doteq M(y)dy
\]
captures the statistics of natural images: It is small almost everywhere, except near image boundaries, where it approaches an impulse. A normalized version of (\ref{eq-h}) yields
\be
\boxed{\bar h(x,\theta) \doteq \frac{h(x,\theta)}{\int_{\mathbb S^1} h(x, \theta)d\theta}} ~~~ {\rm DOG}
\label{eq-nhog}
\ee
Given a (``training'') image $I$, since we cannot determine $\h, w$, we can assume them to be the identity so $I = \rho$, given which the likelihood of a (``test'') image having gradient orientation $\theta$ at $x$ is given by 
\be
p(\theta | \rho(x)) = {\cal N}_\epsilon\left(\theta - \angle G(x) \right)
\ee
and HOG (\ref{eq-h}) is the (un-normalized) density in the variable $\theta \in {\mathbb S}^1$
\be
h(x, \theta)= \int p(\theta | \rho(y)) {\cal N}_\sigma(x-y)\mu(dy)
\label{eq-hog-density}
\ee
and DOG its normalized version.
\cancut{One then can, given a descriptor $h$ constructed from $\rho$, infer the most likely image $\hat I$ up to a contrast transformation point-wise, via
\be
{
\hat G(x) \doteq \arg\max_\theta \bar h(x, \theta)
~~~ {\rm and} ~~~
\hat M(x) \doteq \max_\theta \bar h(x, \theta)}
\label{eq-Ghat}
\ee
then solving a boundary-value problem to integrate $\nabla \hat I(x) \doteq \hat M(x) \hat G(x)$ into an estimate $\hat I(x)$ (Fig. \ref{fig-Ghat}).
\begin{figure}[htb]
\begin{center}
\includegraphics[height=.09\textwidth,width=.09\textwidth]{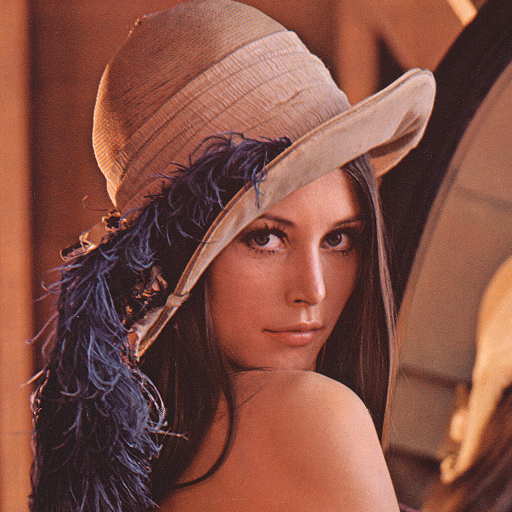}
\includegraphics[height=.09\textwidth,width=4cm]{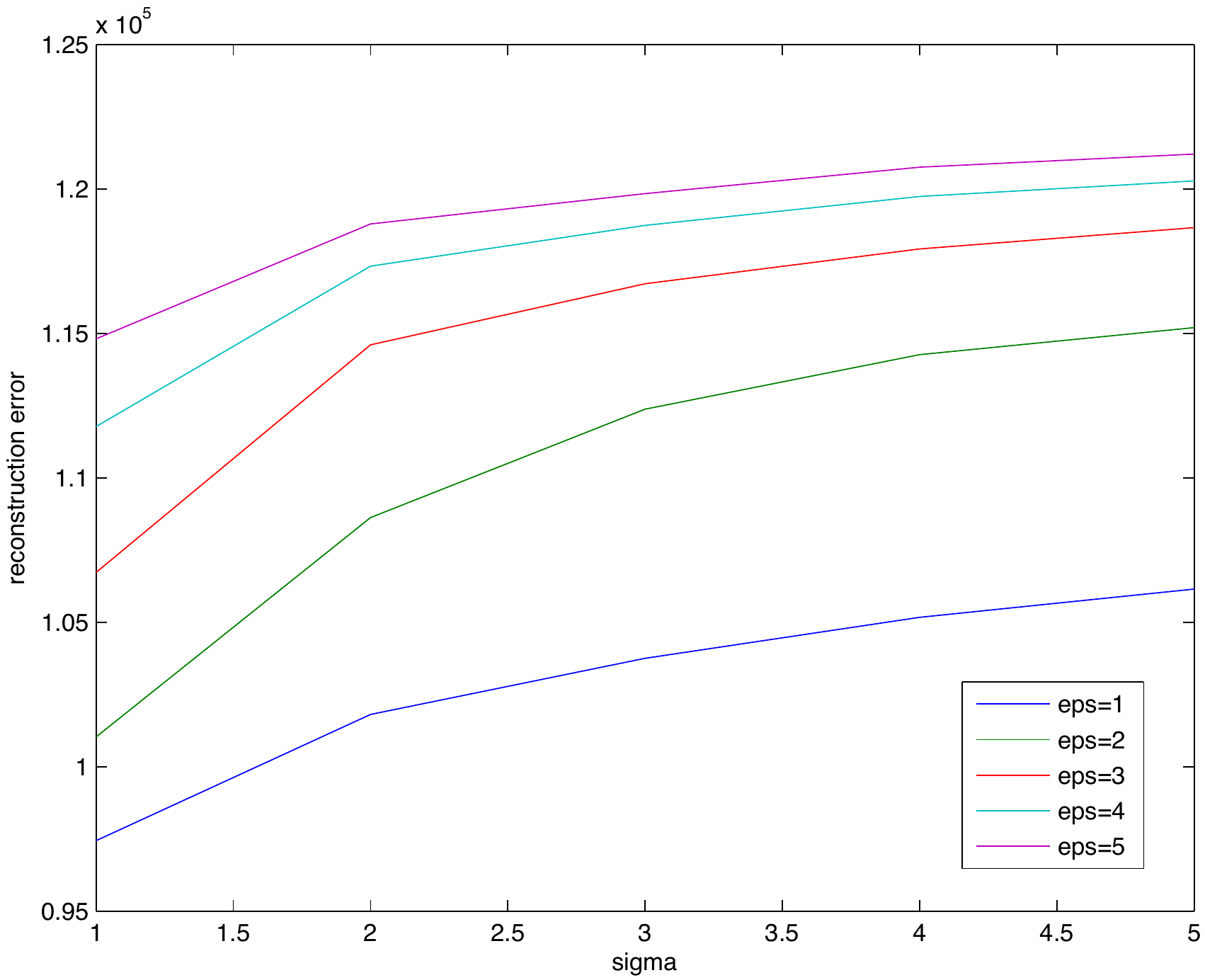}\\
\includegraphics[height=.09\textwidth,width=.09\textwidth]{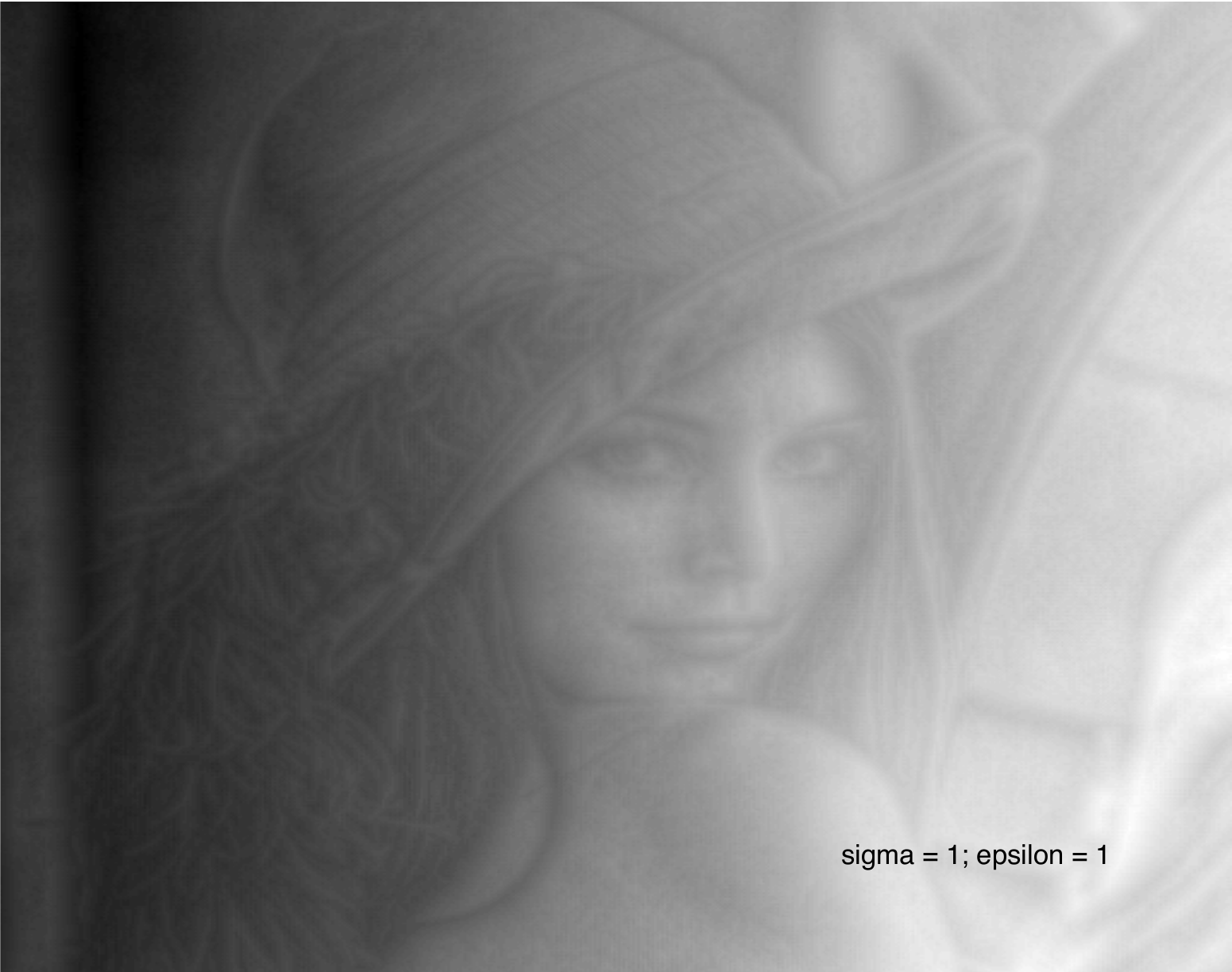}
\includegraphics[height=.09\textwidth,width=.09\textwidth]{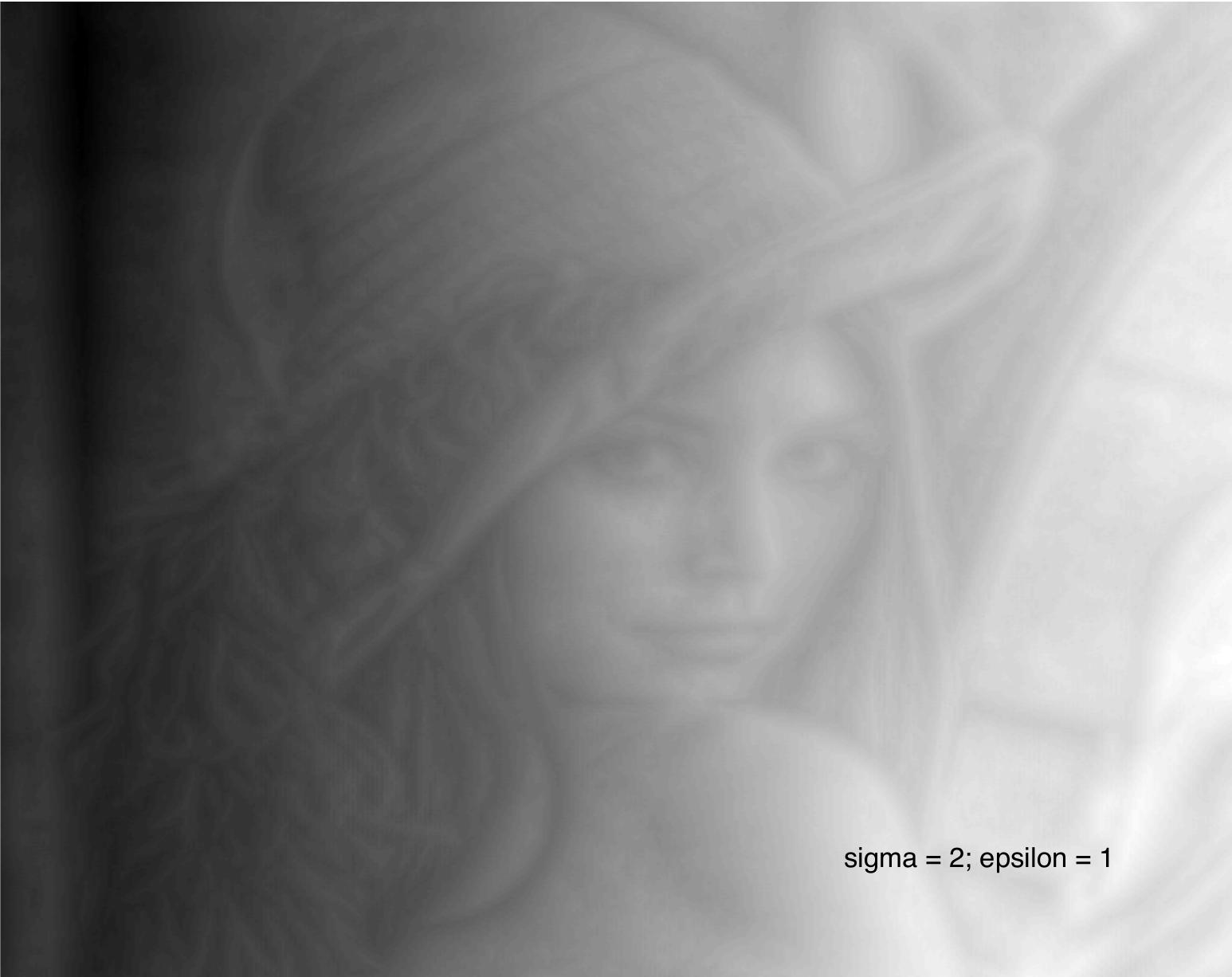}
\includegraphics[height=.09\textwidth,width=.09\textwidth]{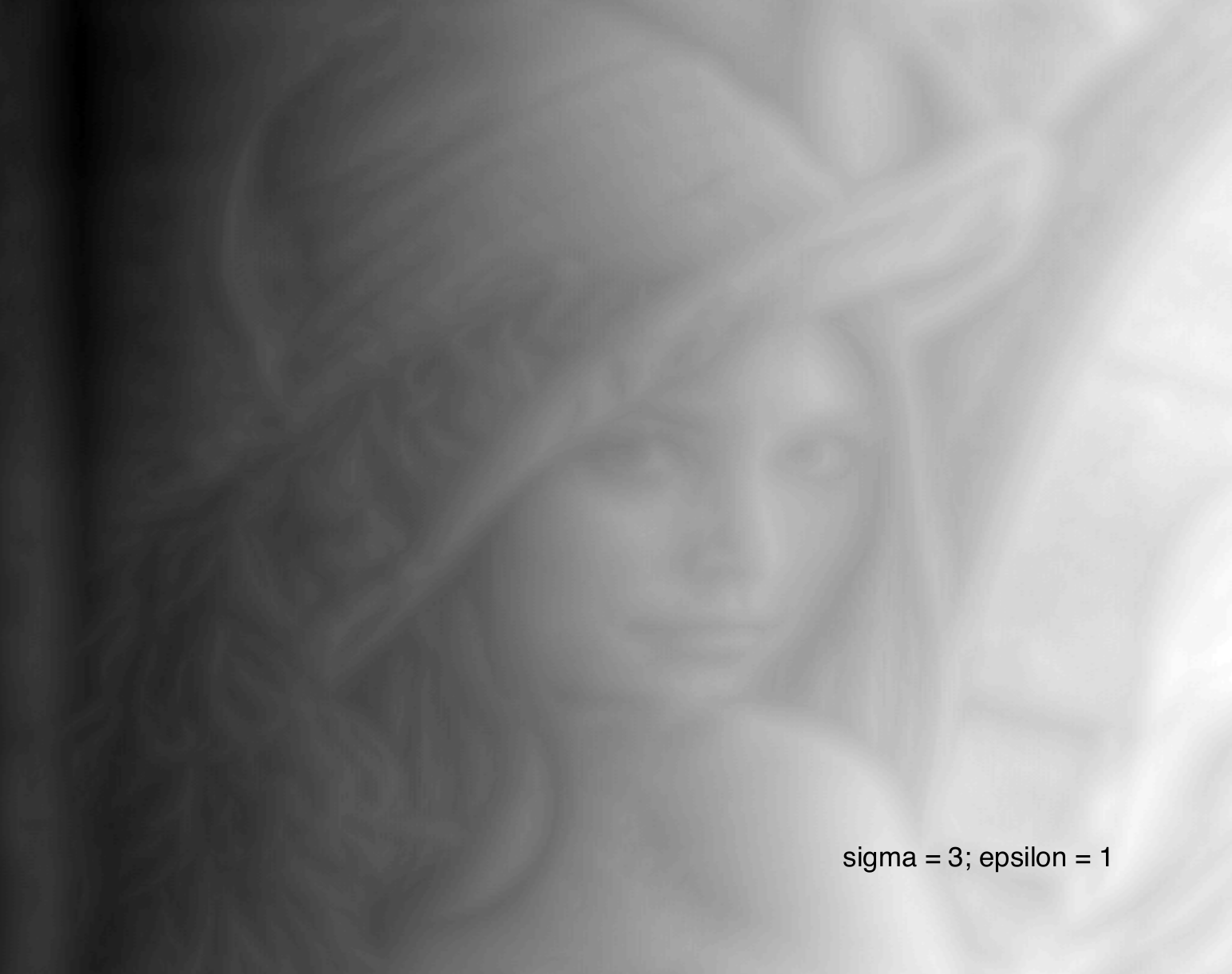}
\includegraphics[height=.09\textwidth,width=.09\textwidth]{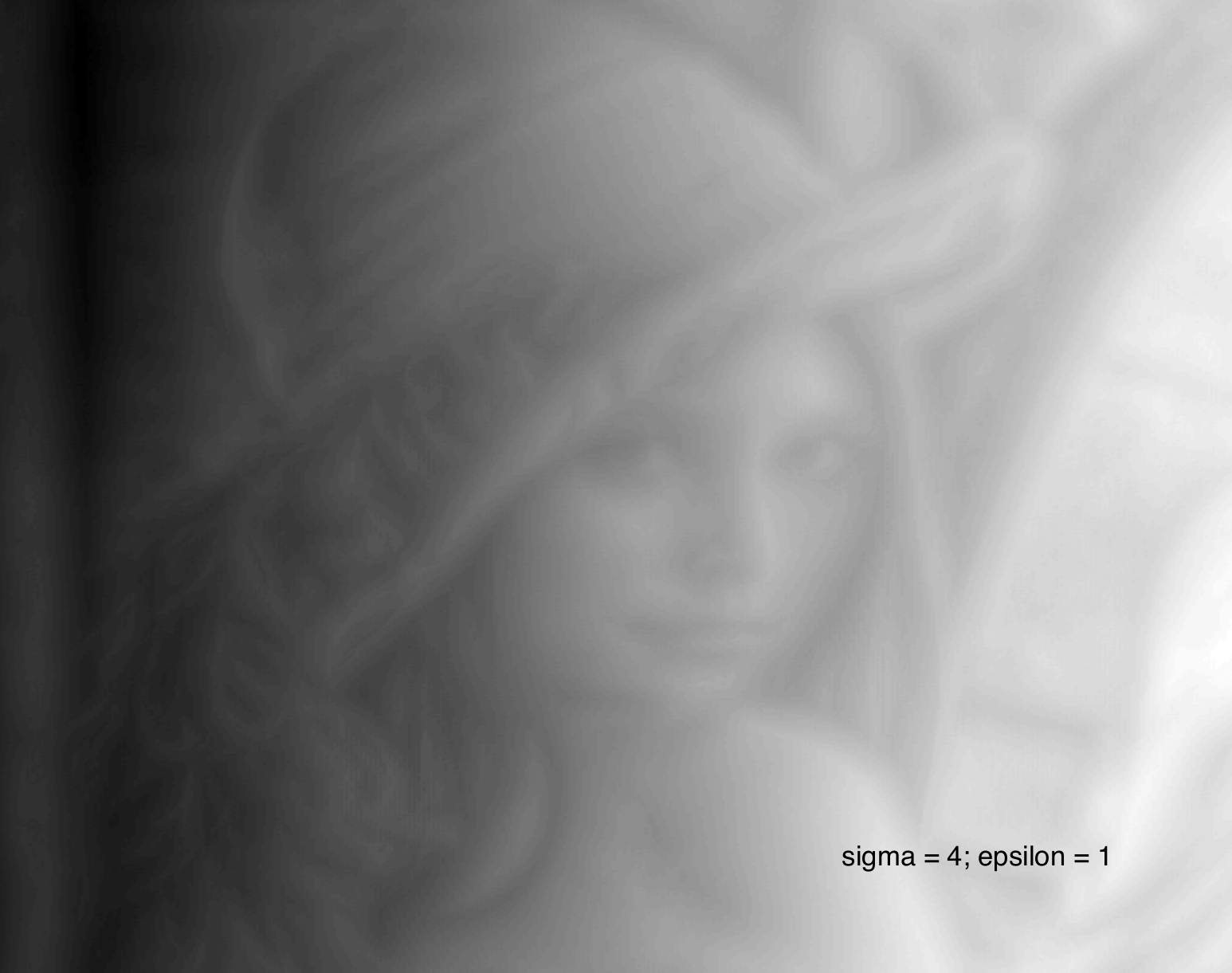}
\includegraphics[height=.09\textwidth,width=.09\textwidth]{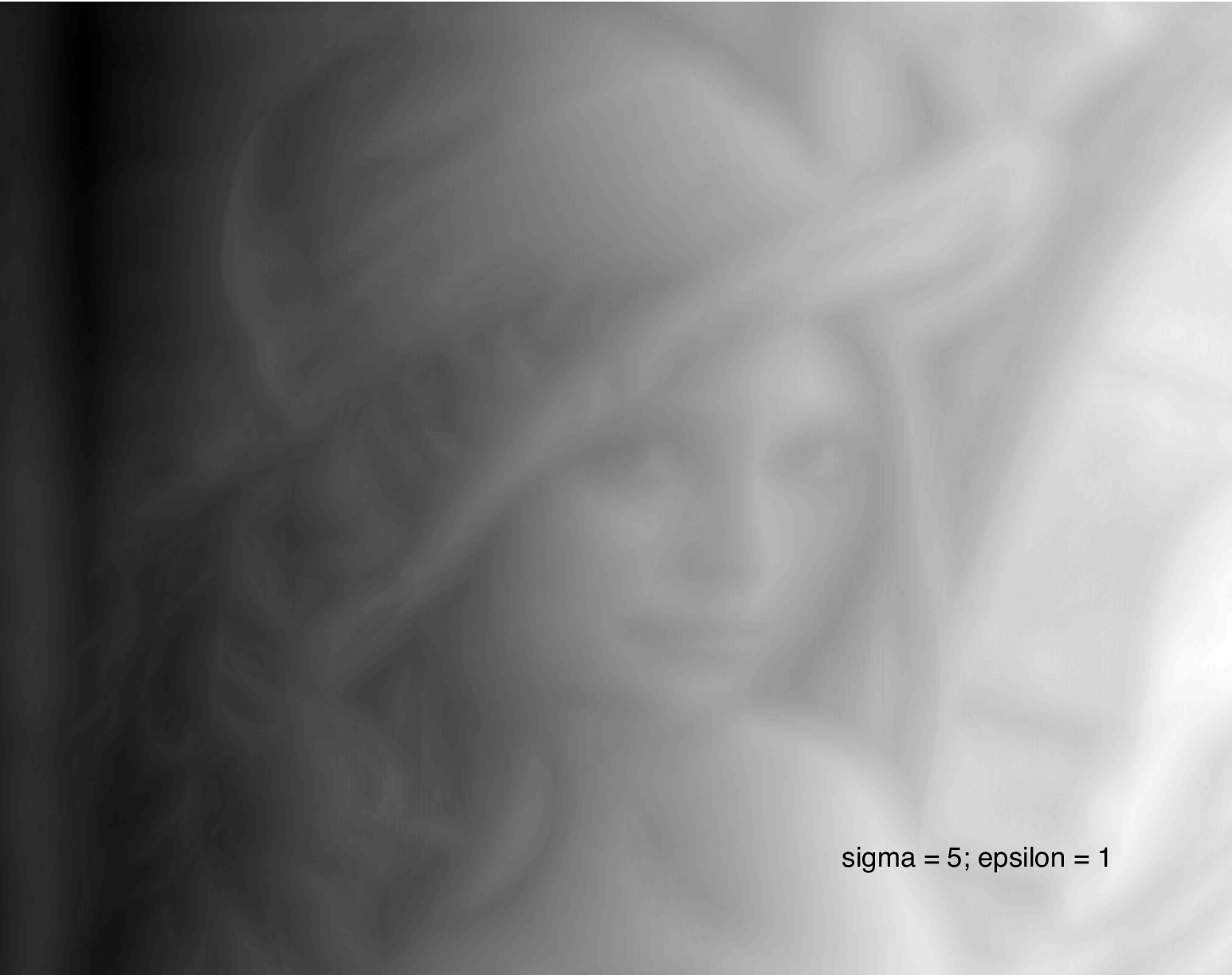}\\
 \includegraphics[height=.09\textwidth,width=.09\textwidth]{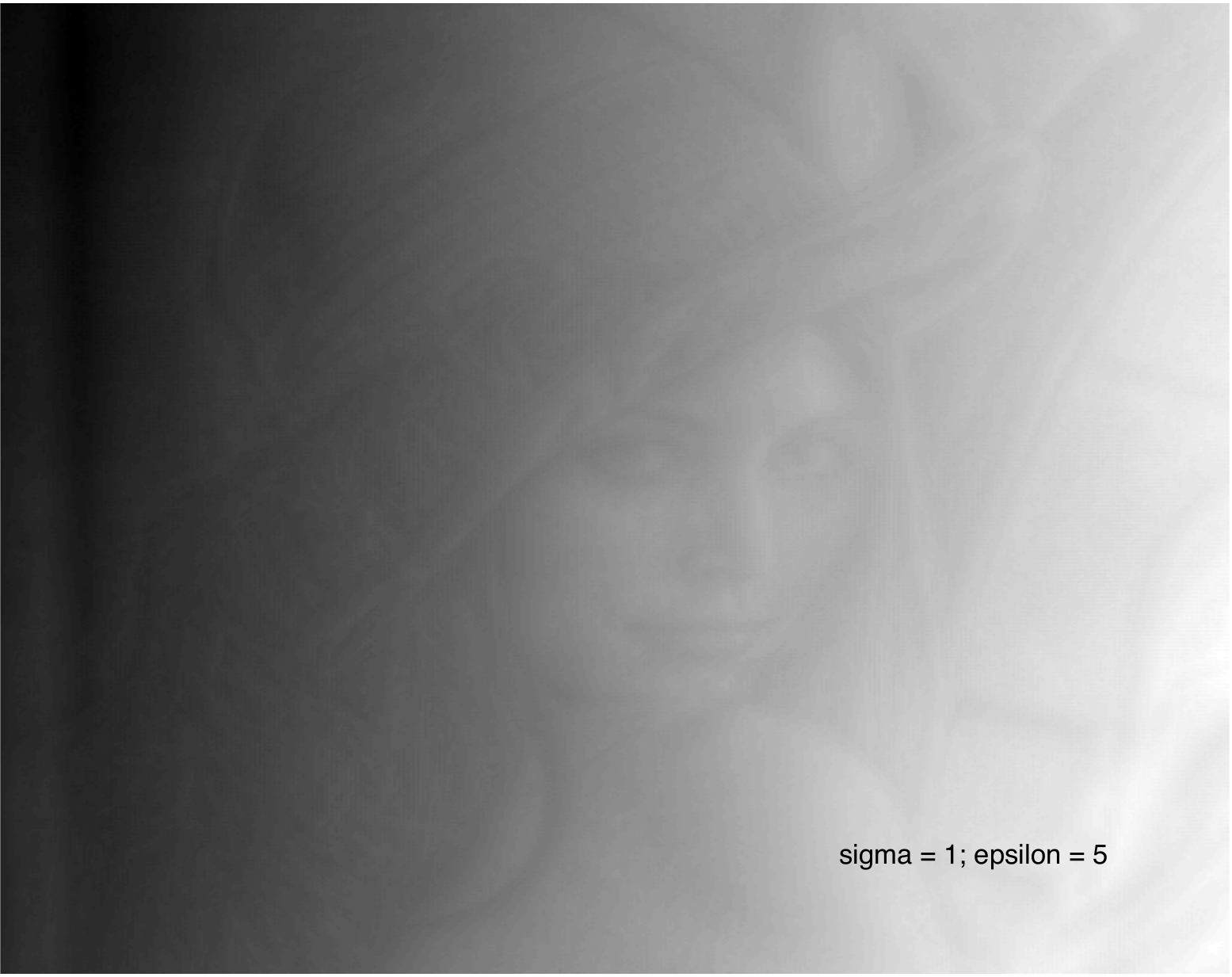}
 \includegraphics[height=.09\textwidth,width=.09\textwidth]{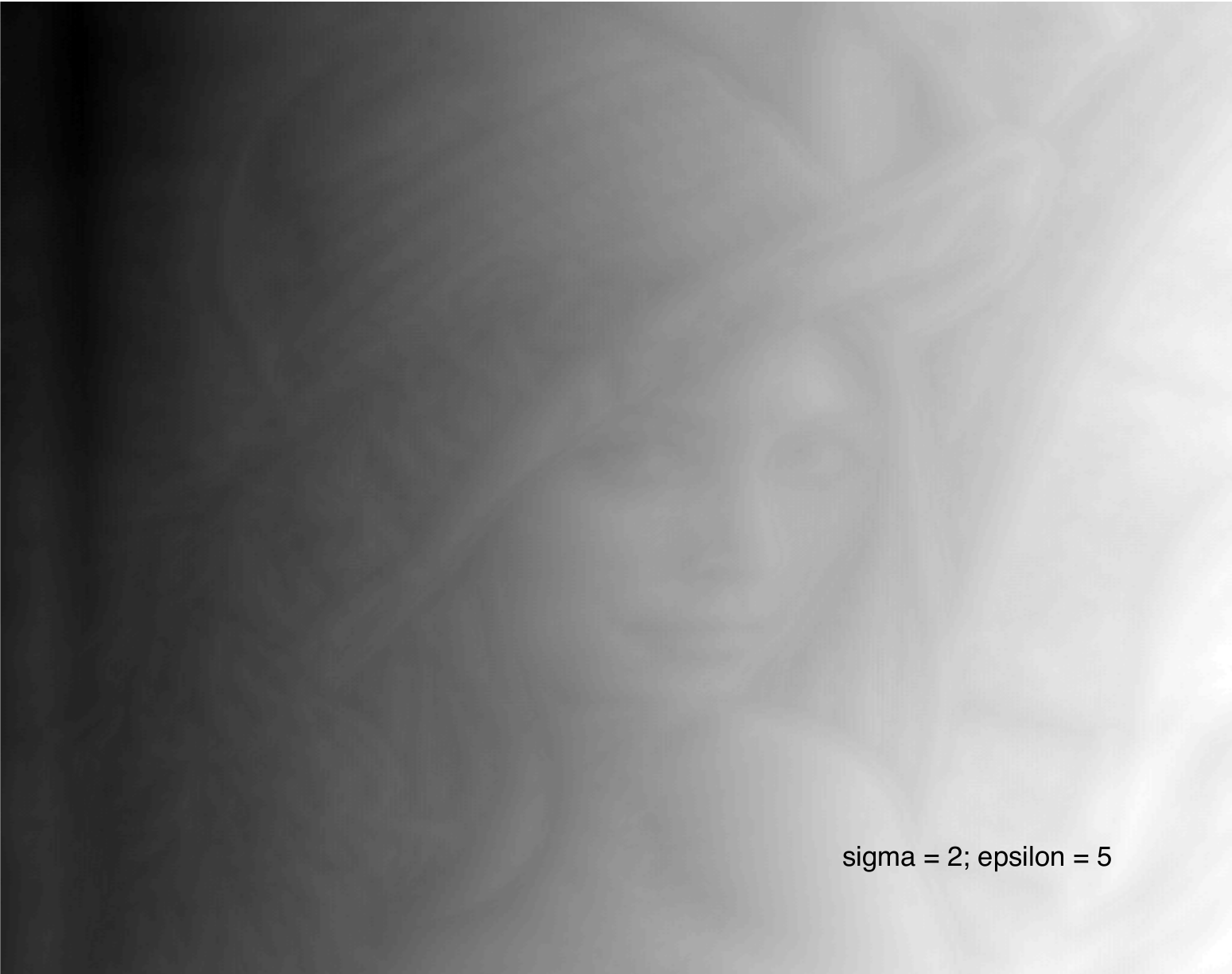}
 \includegraphics[height=.09\textwidth,width=.09\textwidth]{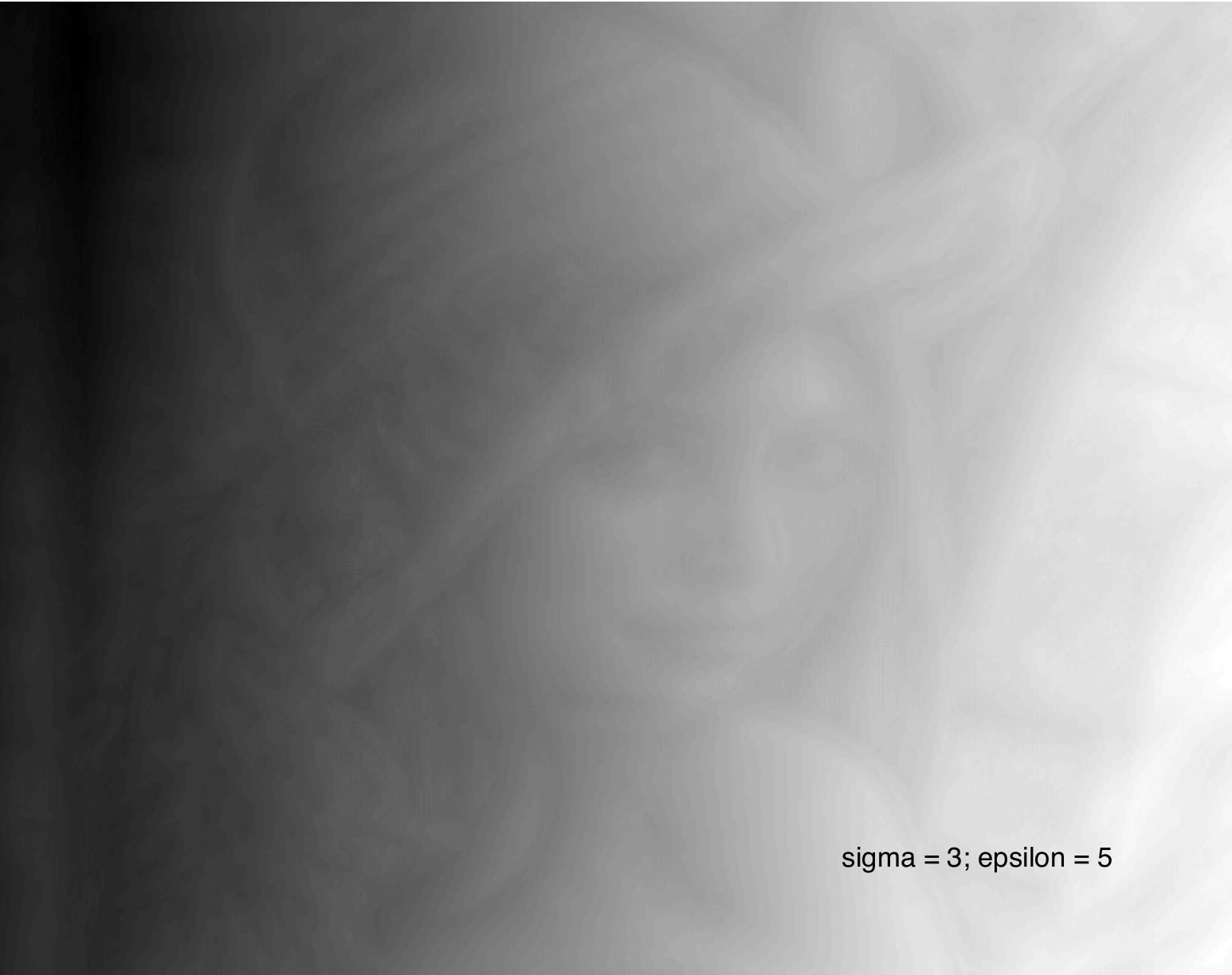}
 \includegraphics[height=.09\textwidth,width=.09\textwidth]{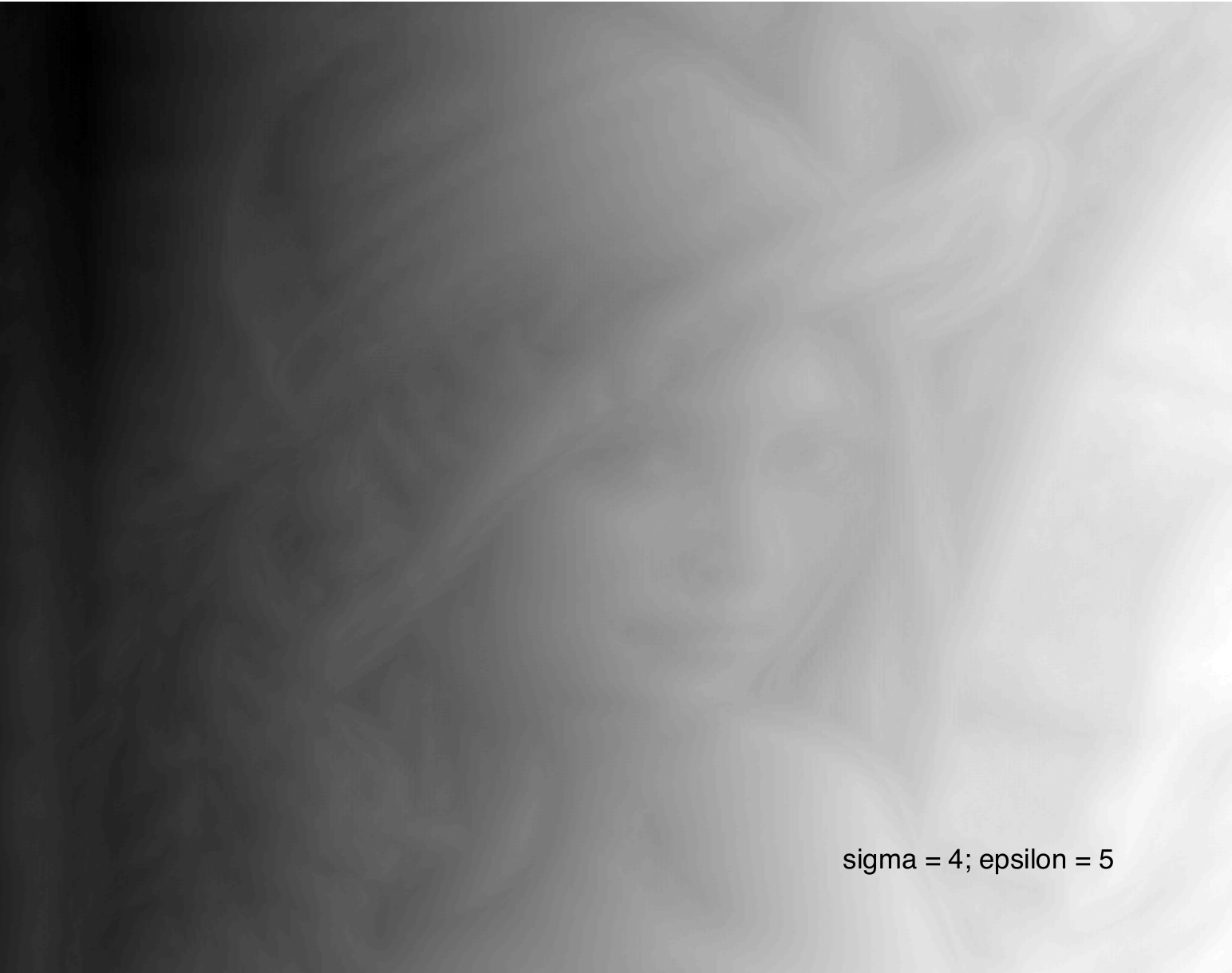}
 \includegraphics[height=.09\textwidth,width=.09\textwidth]{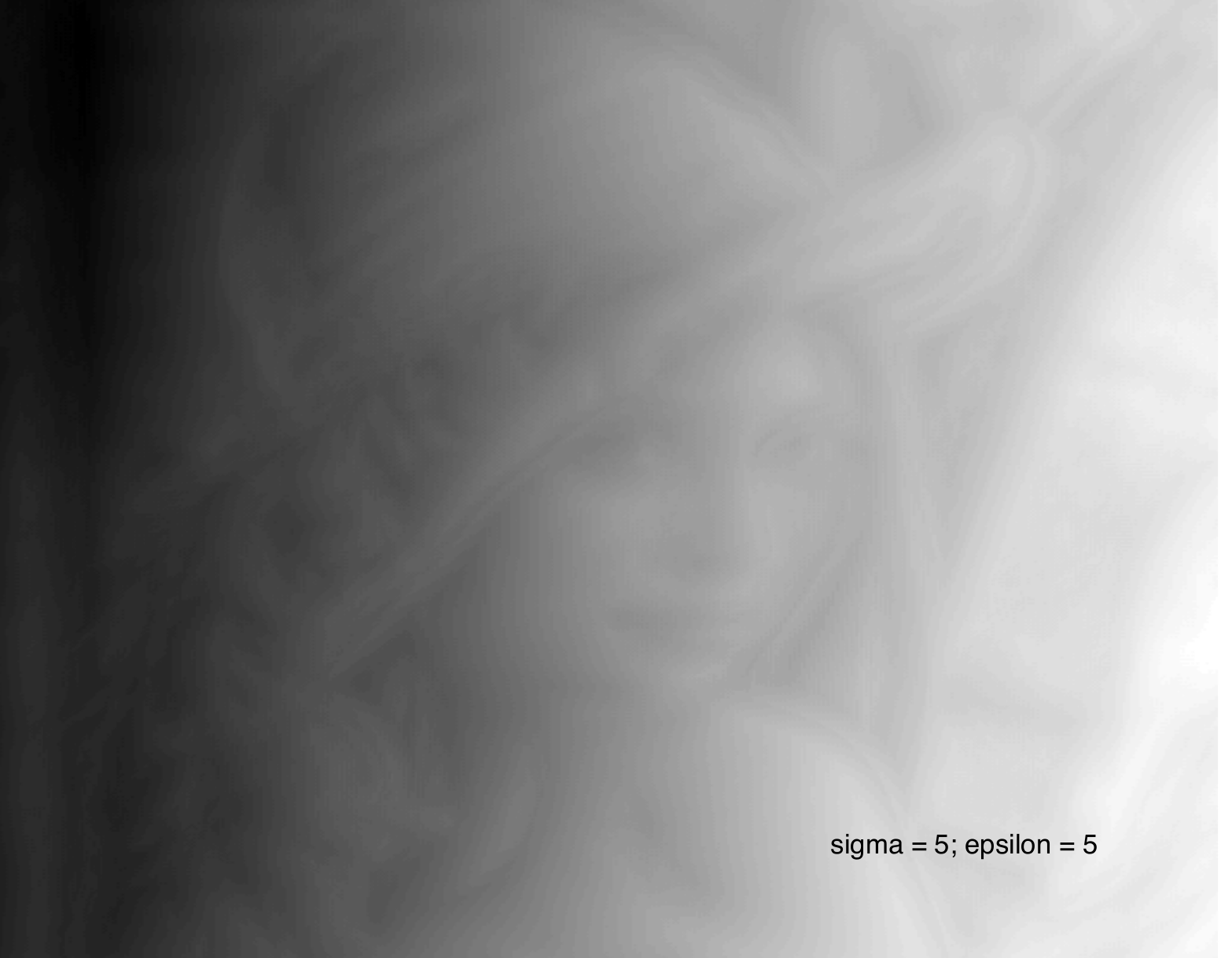}
\end{center}
\caption{\sl HOG inversion: The classic Lena image is used to construct single-view descriptors at varying values of $\sigma, \epsilon$; these are in turn used to reconstruct the gradient field, and integrated to yield the reconstructed images.}
\label{fig-Ghat}
\end{figure}
Since this inversion formula is point-wise, it only provides a reconstruction where the descriptor is computed. A dense reconstruction can be obtained from sparsely sampled descriptors as shown in the appendix. 
}

\subsection{Handling nuisance variability in HOG/DOG}

If one interprets HOG/DOG as a probability density (\ref{eq-hog-density}), it is worth understanding how it relates to the ideal descriptor (\ref{eq-marg}). In particular, {\em What measure $dP(\h_t, w_t)$ does HOG/DOG use to marginalize nuisances?}

\cut{\subsubsection{Contrast management (one view)}}
\label{sect-canonization}

Contrast (monotonic continuous transformations of the range of the image) $\h$, often used to model gross illumination changes, can be eliminated at the outset without marginalization. Despite being simplistic (they do not account for specularity, translucency, inter-reflection etc.), they are an infinite-dimensional group, so defining a base measure, learning a distribution $dP(\h)$, and marginalizing (\ref{eq-marg}) is problematic.

However, contrast transformations do not depend on intrinsic properties of the scene. Neglecting (spatial and range) quantization, the geometry of the level curves of the image is a complete contrast invariant \cite{sundaramoorthiPVS09}, in the sense that it enables reconstructing an image that is equivalent to the original but for a contrast transformation \cite{alvarezGLM93}. Since the gradient orientation $G(x)$ is everywhere orthogonal to the level curves, replacing the intensity-based kernel ${\cal N}_\epsilon(I - \h \circ \rho)$ with a gradient-orientation kernel \cite{watson1983statistics}\footnote{We are overloading the notation by using the same symbol ${\cal N}_\epsilon(\theta -\mu)$ with $\mu \in \real$ to indicate an ordinary univariate Gaussian density, and ${\cal N}_\epsilon(\theta - \mu)$ with $\mu \in {\mathbb S}^1$ to indicate an angular Gaussian.}  ${\cal N}_{\epsilon}(\theta - \angle G(x))$ with $G(x) = \nabla \rho/ \| \nabla \rho \|$, we can annihilate the effects of contrast changes at no loss of discriminative power \cite{sundaramoorthiPVS09}. Following (\ref{eq-hog-density}), {\em given} the radiance $\rho$, and therefore its gradient orientation, $G$,  $p(\theta | \rho(x))$ is either the likelihood of an the albedo having orientation $\theta \in {\mathbb S}^1$, or the likelihood of it having (possibly normalized) intensity value $\theta \in [0, \ 1]$ at $x$. Then (\ref{eq-marg}) at some $x\in D$ is written as 
\be
p(\theta | \rho, S, x) = \int p(\theta | \rho, w,x)dP(w|S) 
\ee
with $p(\theta | \rho, w, x) \doteq {\cal N}_\epsilon(\theta - \angle G \circ w (x))$ and $G = \nabla \rho / \| \nabla \rho \|$. This corresponds to HOG (\ref{eq-hog-density}) if we restrict the nuisance variability to {\em constant} domain translations, $w(x) = x + v$, {\em independent of $S$}: 
\be
dP(w |S) = dP(v) = {\cal N}_\sigma(v)\mu(dv)
\ee
after the change of variables $y = w(x)$ and $v = y-x$ and {\em normalization.} The latter is necessary to obtain a probability density, and also to achieve contrast invariance: Unlike DOG, HOG is in fact {\em not} contrast invariant, as one can easily verify. 
\cut{
\begin{claim}[Contrast management in HOG/DOG]
\label{claim-contrast}
The gradient orientation histogram HOG, weighted by the gradient norm, is {\em not} contrast invariant. However, its normalized version DOG is insensitive to contrast changes up to sampling effects. 
\end{claim}
As an alternative to weighting the gradient direction with its norm, we can use a smooth function that approaches the gradient direction when the norm is large, and approaches zero when it is small, for instance
\be
G(x) = \nabla \rho \frac{\|\nabla \rho\|}{\epsilon + \| \nabla \rho \|^2}
\ee
with some $\epsilon > 0$. This, together with other forms of contrast management described in the next section, has been explored empirically in Sect. \ref{sect-expm} and App. \ref{app-expm}.
}
\begin{claim}[HOG as a (poor approximation of the) class-conditional distribution]
\label{claim-bad}
HOG/DOG (\ref{eq-nhog}) is an approximation of the class-conditional distribution (\ref{eq-marg}) of images of the same underlying scene, where only viewpoint changes that correspond to planar translations {\em of the image} are marginalized. 
\end{claim}
\cut{\subsubsection{Canonization via co-variant detection}}

\cut{In Sect. \ref{sect-canonization} we have replaced $I$ and $\rho$ with a {\em maximal invariant} to contrast transformations, thus eliminating the nuisance and nothing else (to first approximation). When such nuisances form a {\em group}, there is a constructive procedure for doing so, by designing a (co-variant detector) functional that maps the data (\eg an image) onto a ``canonical'' element of the group (\eg contrast transformations), which in turn defines a reference frame relative to which the data is, by construction, invariant to the nuisance group\cut{ \cite{soatto10}}. For the case of affine contrast transformations, a trivial co-variant detector is the one that returns the value of the brightest and the darkest pixels (or the mean and standard deviation of the pixel values); these can then be assigned to zero and one, respectively, after which affine contrast changes have no effect. This is equivalent to pre-processing the image by subtracting the mean (which has therefore value zero henceforth) and dividing by the standard deviation (which has therefore value one henceforth). Such ``centering'' of the data can also be applied in log-space, where linear contrast changes are normalized by dividing the image by its mean, a procedure called ``weberization'' following Weber's laws of perception. All these approaches to contrast management are explored empirically in Sect. \ref{sect-expm}.}
Similarly to contrast, we can eliminate small-dimensional group transformations of the domain at the outset, though a {\em co-variant detector}, that determines a local reference frame that co-varies with the group, and therefore the data in such a moving frame is, by construction, invariant to it. Because co-variant detectors do not commute with non-invertible transformations such as occlusion or spatial quantization, it can be shown \cite{soatto10} that one has to restrict the attention to {\em local neighborhoods} of co-variant detector frames. 
The literature offers a variety of location-scale co-variant detectors, \eg extrema (in location and scale) of the Hessian of Gaussian, Laplacian of Gaussian, or Difference of Gaussian convolutions of the data, in addition to other ``corner detectors'' applied to any scale-space.\cut{ In Sect. \ref{sect-expm} we choose \cite{fast}, computed on the scale-space, for its simplicity.} Planar rotation can also be eliminated using the maximal gradient direction \cite{lowe04distinctive}, or based on an inertial reference (\eg, the projection of gravity onto the image plane\cancut{ \cite{jonesS09}}), if the scene of interest is geo-referenced.
\cut{Canonization can also be performed on planar domain deformations. It can be shown\cut{ in \cite{soatto10}} that this can be done in a lossless fashion (to first order) only for planar similarity transformations (translations, rotations, scaling). Because co-variant detectors do not commute with non-invertible transformations (scaling/quantization and occlusion), one cannot choose {\em one} canonical representative, but instead all local extrema of the co-variant detector are retained as putative canonical elements. The literature already offers a variety of translation-scale co-variant detectors, for instance the extrema (in location and scale) of the Hessian of Gaussian, Laplacian of Gaussian, or Difference of Gaussian convolutions of the data, in addition to other ``corner detectors'' applied to any scale-space. In Sect. \ref{sect-expm} we choose \cite{fast} for its simplicity. Rotation can be canonized based on a photometric reference, for instance the direction that maximizes the norm of the gradient at a point, as in \cite{lowe04distinctive}, or better based on an inertial reference (\eg, the projection of gravity onto the image plane \cite{jonesS09}), if the scene of interest is geo-referenced.} Therefore, from now on we assume planar similarities are removed, and deal with the residual domain deformations.  Even so, however, marginalization in HOG is relative to a distribution that is independent of $S$:
\begin{claim}[HOG is not shape-discriminative]
\label{claim-no-shape}
The HOG/DOG descriptor (\ref{eq-nhog}) is agnostic of changes in the shape of the underlying scene that keep the (one) image constant.
\end{claim}
One should not confuse the shape of the {\em scene} $S\subset \real^3$, which depends on its three-dimensional geometry, with the (two-dimensional) shape of the intensity profile of the {\em image}, which is a photometric property and is encoded in $\rho$. Clearly HOG depends on the latter, but it is insensitive to any deformation of the scene that would yield the same projection onto the (single) image. The two are related only at occluding boundaries, that however are excluded from our analysis, as well as from the typical use of local descriptors (although see \cite{vedaldiS06ECCV,ayvaciS11}).

\section{Extensions to multiple views}

 In a sequence of images, consistent co-variant frames can be determined by a {\em tracker} (Sec. \ref{sect-expm}) and eliminated at the outset with contrast transformations as in Sec. \ref{sect-canonization}. The result is a sequence of (similarity-invariant) images/patches, that sample the residual nuisance variability, as by assumption they all portray the same underlying scene $\rho, S$.

\subsection{Extension via Sampling}
\label{sect-mv-hog}

{\em If we were given $\rho$,} multiple views $\{I_t\}_{t = 1}^T$ would provide us with samples $\{ w_t\}_{t=1}^T$ from the (class-specific) distribution $dP(w|S)$, via dense correspondence $I_t = \rho \circ w_t$ (or optical flow, $I_t(x) = \rho(x+v_t(x))$). Ideally, one would use the samples to compute a Parzen-like estimate of the distribution, but this presents technical challenges due to the curse of dimensionality. However, {\em one could use the (space-varying) samples $v_t(x)$ to define a (space-varying) measure $d\hat P(v,x|S)$ on {\em constant} displacement fields} $v \in \real^2$, via
\be
d\hat P(v, x | S) = \frac{1}{T}\sum_{t=1}^T {\cal N}_\sigma(v - v_t(x))\mu(dv) 
\ee
where the dependency on $S$ is through the samples $v_t(x)$. Using this measure,\cancut{\footnote{Note that this is not a proper measure on the group of diffeomorphisms.}} and calling $y \doteq w(x) = x+v$, so $dy = dv$, under {\em sufficient excitation conditions} \cite{bitmead84} we can approximate $p(\theta | \rho, S)$ in (\ref{eq-marg}) with 
\bea
p(\theta | x, \rho, S)  &\simeq& p(\theta | x, \rho, \{w_t\}) = \frac{1}{T}\sum_{t=1}^T \int {\cal N}_\epsilon(\theta - \angle G(y)) d\hat P(y-x, x | S) = \nonumber \\ 
&= &
 \frac{1}{T}\sum_{t=1}^T  \int p(\theta | \rho, y) {\cal N}_\sigma(y-w_t(x))\mu(dy)
\eea
This could be considered an extension of HOG/DOG to multiple views, {\em if we knew $\rho$}. Unfortunately, we only measure $I_t$, related to $\rho$ via (\ref{eq-I}). A further approximation can be made by assuming that $v_t(x)$ is smooth, so the Jacobian $J_v(x)$ is small, and $v_t(y) = v_t(x+v(x)) \simeq v_t(x) + J_v(x+v(x))v(x) \simeq v_t(x)$. Applying the change of variable $v-v_t(x) \rightarrow y - x$, (and since $x$ and $v_t$ are held constant, $dv = dy$), we obtain 
\bea
p(\theta | x, \rho, S)  & \simeq & \frac{1}{T}\sum_t \int {\cal N}_\epsilon(\theta - \angle G(x+v)) {\cal N}_\sigma(x + v - x - v_t(x))\mu(dv) = \nonumber \\ 
&=&
\frac{1}{T}\sum_t \int p(\theta | I_t(y)) {\cal N}_\sigma(y - x)\mu(dy).
\label{eq-h-t}
\eea
since $I_t(y) = \rho \circ w_t(y) = \rho(y + v_t(y)) + n_t(y)$, and the effects of the approximation $v_t(y) \simeq v_t(x)$ can be absorbed by inflating the noise covariance $\epsilon$.\cancut{\footnote{ An alternate derivation can be obtained via Monte Carlo approximation; while the group of diffeomorphisms is infinite-dimensional, formally we can write: 
\[
dP(I|\rho, S) = \int {\cal N}_\epsilon (I-\rho \circ w)dP(w|S) \simeq \sum_{w_t \sim dP(w|S)} {\cal N}_\epsilon(I - \rho \circ w_t) = \sum_{t} {\cal N}_\epsilon(I - I_t) 
\]
Unfortunately, $\{w_t\}_{t=1}^T$ cannot be a (sufficiently exciting) {\em fair sample} from $dP(w|S)$. To introduce some regularization, and to take into account residual translations from the tracker, one can perform spatial blurring and enforce the statistics of natural images, thereby obtaining
\[
dP(I(x)|\rho, S) \simeq  \sum_{t} \int {\cal N}_\epsilon(I(x) - I_t(y)){\cal N}_\sigma(x-y)\mu(dy) 
\]
which is the same expression as (\ref{eq-h-t}).}} This can be easily implemented given a collection of images $\{ I_t\}_{t=1}^T$, and can be used to compute the likelihood of a test image $I$, or can be interpreted as the class-conditional distribution $p(I | \{ I_t\}) \simeq p(I | \rho, S)$ in (\ref{eq-marg}), assuming the sample $\{ I_t\}$ is {\em sufficiently exciting}: 
\cut{\begin{claim}[MV-HOG]
A multi-view extension of HOG can be obtained by approximating the nuisance distribution with a (shape-specific) space-varying measure learned from sample views of the same scene.}
Writing it explicitly as a function of location $x\in D$, angle $\theta \in {\mathbb S}^1$, given a collection of images (patches) $\{ I_t\}_{t=1}^T$, we have
\be
\boxed{h_{\rm MV}(x,\theta) = \frac{1}{T}\sum_t \int_{\real^2} {\cal N}_{\epsilon} \left( \theta - \angle{\frac{\nabla I_t(y)}{\| \nabla I_t\|}} \right) {\cal N}_\sigma \left( x - y \right) \| \nabla I_t(y) \|  dy} 
\label{eq-h-tbox}
\ee
which we call MV-HOG. Its contrast-insensitive version $\bar h(x,\theta)$ is obtained by normalization, as usual. We then have
\be
\bar h_{\rm MV}(x, \theta) = p(\theta | x, \{I_t\}) \stackrel{{\rm suff. \ exc.}}{\longrightarrow} p(\theta | x, S, \rho)
\ee
asymptotically as $T \rightarrow \infty$ under sufficient excitation conditions.

\subsection{Extension via reconstruction}
\label{sect-marginalization}
{
To compute a better marginalization, given {\em more than one} image of the same scene, we can perform (dense) reconstruction of shape $\hat S$, radiance $\hat \rho$ and motion $\hat g_t \in SE(3)$ for instance using \cite{faugerasK96,jinSY05IJCV,Graber2011}, by solving 
\[
\hat S, \hat \rho, \hat g_t = \arg\min_{S, \rho, g_t} \sum_t \int_D | I_t(x) - \rho \circ \pi\circ g_t^{-1}\circ \pi_S^{-1}(x)|^2 dx + \lambda \int_S dA \]
where surface area of $S$ is used as a regularizer\cancut{ (a regularizer for $\rho$ is usually not needed under the Lambertian assumption \cite{soattoY02cvpr})}. This yields\footnote{Dense reconstruction is a difficult (non-convex, ill-posed, infinite-dimensional) optimization problem, that often fails to yield accurate solution when the data is not ``sufficiently exciting.'' For instance, multiple images of a white scene does not enable reconstructing ints shape. However, in our context, when conditions prevent us from accurately recovering shape, it means that any deformation is equally likely. So the resulting descriptor is already invariant to nuisance deformations.} a point estimate of shape $\hat S$, from which an object-specific measure $dP(w|\hat S)$ can be constructed using a base measure on $SE(3)$, $d\mu(g)$: 
\be
dP(w|\hat S) =  \int_{SE(3)} p(w | g, \hat S)dP(g) =   \int_{SO(3)} \delta(w - \pi\circ g \circ \pi_{\hat S}^{-1}) d\mu(g). \nonumber
\ee
$d\mu(g)$ is uniform on $SO(3)$. The restriction from $SE(3)$ to $SO(3)$ is possible thanks to the elimination of planar similarities, leaving us with marginalizing on a compact group with a proper Haar measure. If orientation is fixed by gravity, the measure can be further restricted to rotations about gravity $\mathbb S^1$.\cancut{
 Note that this measure is not defined on the entire group of diffeomorphisms, $dP(w)$,  but only those generated by moving around an object with shape $\hat S$, $dP(w|S)$.\footnote{If $l \doteq \pi\circ g \circ \pi_{\hat S}^{-1}$ a measure $\mu$ on $SE(3)$ induces a measure on diffeomorphisms restricted to $S$ via the push-forward $l_*(\mu)$.}} From this a more specific descriptor can be constructed, which we call R-HOG:

\be
\boxed{h_{\rm R}(x, \theta) = \int p(\theta | \hat \rho, \hat S, w) dP(w| \hat S) = \int_{SO(3)} {\cal N}_\epsilon(\theta  - \angle G_{\hat \rho}\circ \pi\circ g \circ \pi_{\hat S}^{-1}) d\mu(g)}
\label{eq-margin}
\ee

\noindent where $G_\rho \doteq \nabla \rho / \| \nabla \rho \|$. As usual, we achieve insensitivity to contrast by normalization, obtaining $\bar h_{\rm R}(x, \theta)$, and
\be
\bar h_{\rm R}(x, \theta) = p(\theta | x, \hat \rho, \hat S) \stackrel{{\rm suff. \ exc.}}{\longrightarrow} p(\theta | x, S, \rho)
\ee
asymptotically if the estimator $\hat \rho, \hat S$ is unbiased. Note that informationally, $\{\hat \rho, \hat S\}$ and $\{I_t\}_{t=1}^T$ are equivalent under sufficient excitation, but computationally they embody very different philosophies (Monte Carlo for MV-HOG vs. Maximum Likelihood for R-HOG), and yield different implementations. Both MV-HOG and R-HOG have the same complexity of HOG, and can be compared as we describe next.

\section{Evaluation and comparison of descriptors}
\label{sect-comparison}

The descriptors (\ref{eq-h-tbox}) or (\ref{eq-margin}) can be interpreted as statistics,  deterministic functions of the {\em training set}. In fact, we have 
\be
h(x,\theta) = {\mathbb E}_q\left( p(\theta | x, \rho, S) \right)
\ee
where the expectation is taken with respect to $q = {\cal N}_\epsilon(x-y)\mu(dy)$ for HOG, $q = \sum_t {\cal N}_\epsilon(x-w_t(y))\mu(dy)$ for MV-HOG, and $q = \delta(\rho-\hat \rho)\int \delta(w-\pi\circ g \circ \pi^{-1}_{\hat S})d\mu(g)$ for R-HOG.

Once sampled, $h$ can be thought of as a vector having the dimension of the lattice where $x$ is evaluated, times the number of bins where the gradient $\theta$ is quantized. They can then be compared to a descriptor computed from a single view as an element of the ambient linear space (even though they do not live in a linear spaces), for instance the $\ell^1$ norm of the difference, or the correlation coefficient, as customary \cite{lowe04distinctive}.

Alternatively, they can be interpreted as class-conditional densities $p(\theta | \{I_t\}_{t=1}^T)$ for MV-HOG, where presumably $I_t \sim dP(I | S, \rho)$, or $p(\theta | \hat \rho, \hat S)$ for R-HOG, and compared using any distance or divergence between distributions, for instance Kullback-Liebler, Bhattacharyya, $\chi^2$, \etc.
  
Finally, they can be interpreted as likelihood functions; as such, they provide the likelihood that a given test image has gradient orientation $\theta$ at $x$, under the model implied by the training set.

Yet another alternative is to {\em max-out} the nuisance, by evaluating  $p(\theta | \hat \rho, \hat S) = \max_{g\in SE(3)} p(\theta | \hat \rho, \hat S, g) = 
{\cal N}_\epsilon(\theta - \angle G_\rho \circ \pi \circ g^{-1} \circ \pi_{\hat S}^{-1}).$
The above maximum corresponds to the minimum of the least-squares residual
\be
\log p(\theta | \hat \rho, \hat S) = \min_{g\in SE(3)} \int_{SO(3)} | \theta  - \angle G_{\hat \rho} \circ \pi \circ g^{-1} \circ \pi_{\hat S}^{-1}(x) |^2dx.
\label{eq-max}
\ee
Experiments comparing the marginalized descriptor (\ref{eq-margin}), the max-out descriptor (\ref{eq-max}), and the MV-HOG descriptor (\ref{eq-h-tbox}) are reported in Sect. \ref{sect-expm}.

\section{Experiments and Evaluation}
\label{sect-expm}

\subsection{Dataset}
\label{sect-dataset}

\begin{figure}[htb]
\begin{center}
\includegraphics[width=0.95\linewidth]{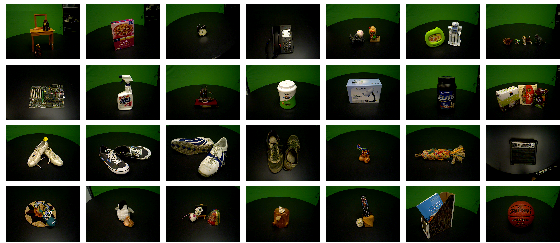}
\end{center}
\caption{{\sl Dataset.} Sample objects in the dataset constructed for evaluation. In addition, synthetic samples are generated by texture-mapping random images onto solid models available in MeshLab.} 
\label{fig-dataset}
\end{figure}
Many datasets are available to test image-to-image matching, where both training and test sets are individual images, each of a different scene. Fewer are available for testing multi-view descriptors \cite{moreels2007evaluation}, and to the best of our knowledge none provide pixel-level correspondence that can be used for validation. Even more problematic, they do not provide a separate test set. We sought objects from  \cite{moreels2007evaluation} to capture a new test set, but those are no longer available. We explored new multi-modal datasets \cite{Geiger2012CVPR}, where ground truth tracking and range are provided, but with no separate test set. We explored using commercial ground imaging, but most are absent in Karlsruhe due to German privacy laws. 


We have therefore constructed a new dataset, similar in spirit to \cite{moreels2007evaluation}, but with a separate test set and dense reconstruction for validation, using a combination of real and synthetic (rendered) objects. The latter are generated by texture-mapping random images onto surface models available in MeshLab. The former are household objects of the kind seen in Fig. \ref{fig-dataset}. Some with significant texture variability, others with little; some with complex shape and topology, others simple. In each case, a sequence of  (training) images per object is obtained moving around the objects in a closed trajectory at varying distance. For real objects, the $400$-frame long trajectory circumnavigates them so as to reveal all visible surfaces; for synthetic ones the $200$ frames span a smaller orbit (Fig. \ref{fig-setup}). 
\begin{figure}[t]
\begin{center}
\includegraphics[height=0.25\textwidth,width=0.85\linewidth]{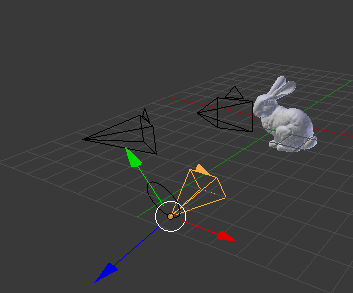}
\end{center}
\caption{{\sl Data Generation.} Synthetic 3D object models are texture-mapped with random city texture (not shown in the modeling snapshot). The trajectory of the camera in the training set is shown as an ellipse attached to the orange frustum. Test images are generated from a camera at vantage points (black frusta) sufficiently far from the training trajectories.}
\label{fig-setup}
\end{figure}

{\noindent \bf Ground Truth:}
We compare descriptors built from the (training) video and test single frames, or descriptors built from them, by first selecting test images where a sufficient co-visible area is present. To establish ground truth, we reconstruct a dense model of each (real) object using an RGB-D (structured light) range sensor and a variant of Kinect Fusion (YAS), for which code is available online. The reconstructed surface enables dense correspondence between co-visible regions in different image by back-projection. This is further validated with standard tools from multiple-view geometry, using epipolar RANSAC bootstrapped with SIFT features. Occlusions are determined using the range map. 

\subsection{Implementation Details}

{\noindent \bf Detection and Tracking:}
Although our goal is to evaluate {\em descriptors}, and therefore we could do away with the detector, one has to select where the descriptor is computed. We use FAST \cite{rosten_2006_machine} as a mechanism to (conservatively) eliminate regions that are expected to have non-discriminative descriptors, but this step could be forgone. Scale changes are handled in a discrete scale-space, \ie images are downsampled by half up to 4 times and FAST is computed at each level. We select different brightness difference thresholds in FAST to ensure roughly the same number of detections in each frame. A minimum distance between features is also enforced to avoid tracking ambiguities; short-baseline correspondence is established with standard MLK \cite{lucasKanade81}. A conservative rejection threshold is chosen to favor long tracks. A sequence of image locations is returned by the tracker for each region, which is then sampled in a rectangular neighborhood at the scale of the detector. We report experiments on two window sizes, $11\times 11$ and $21\times 21$, illustrative of a range of experiments conducted. The sequence of such windows are then used to compute the descriptors. 

{\noindent \bf Descriptors:}
We use HOG from \cite{vlfeat} as baseline, computed on each patch at each frame as determined by the detector and tracker. We set the spatial division parameter to ensure $4\times 4$ cells are created for both patch sizes. The dimension of the resulting HOG is thus $4\times4\times31$. To be consistent with common practice, we perform no post-normalization. 
MV-HOG is implemented according to Sect. \ref{sect-mv-hog}: For each pixel $x$ we build a histogram of gradient orientations at all location $y$'s within a neighborhood ${\cal B}_{2\sigma} (x)$, weighted by the distance from $y$ to the center $x$, and the gradient magnitude. We linearly interpolate orientation bins with cut-off $\epsilon$, as an approximation of the angular Gaussian kernel ${\cal N}_\epsilon$. We choose $2\sigma$ to be a quarter of the patch size so as to have the same spatial weighting of HOG. The number of orientation bins is set to $16$, and temporal aggregation is computed incrementally. Finally, we post-normalize the descriptor according to \eqref{eq-nhog}, thus obtaining a quantized DOG.

{\noindent \bf Reconstruction and marginalized descriptors:}
\label{sect-recon}
To compute an approximation of the marginalized descriptor \eqref{sect-marginalization}, in lieu of a dense 3-D reconstruction from each tracked sequence we computed an approximate dense reconstruction of the entire object, and compared it to the reconstruction obtained from YAS. Since the former had significant artifacts, to obtain a performance upper-bound, we used the restriction of the reconstruction computed from YAS to the pre-image of the tracked patches to build the marginalized descriptors. Using the tracked point as the origin of the local reference frame, we back project the regular lattice around the origin onto the surface, rotate it, and project back onto the image plane to simulate different vantage points $\sim d\mu(g)$. 
We use $10$ keyframes sampled uniformly from the training sequence, and sample a viewing hemisphere with $80$ vantage points. We check the orientation of the surface with the inner product of its normal with the optical axis to ensure visibility. Fig. \ref{fig-recon} shows sample synthesized patches; these are stored in the database. Max-out in Eqn. \eqref{eq-max} amounts to searching through these samples.

\begin{figure}[t]
\begin{center}
\includegraphics[height=0.25\textwidth,width=0.75\linewidth]{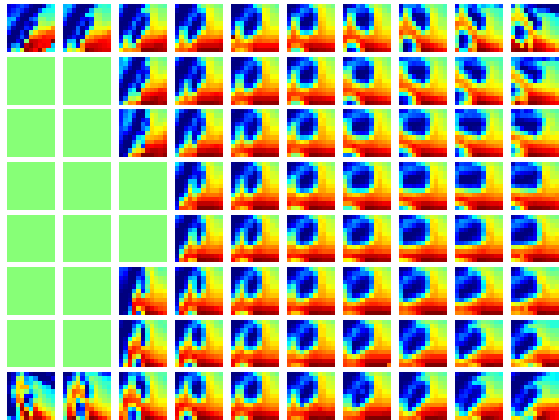}
\end{center}
\caption{{\sl Viewpoint Synthesis.} A local patch (center) is used to synthesize  patches from different vantage points. We restrict the synthesis to $8$ rotations around the $X-Y$ plane and $-\pi/3$ to $\pi/3$ in-plane rotations. A normal test is performed to reject patches (green) that are not visible. The artifacts in the lower-left are caused by the inaccurate estimates of the normal vector.}
\label{fig-recon}
\end{figure}

{\noindent \bf Test descriptors:}
In order to minimize the effects of the detector in the evaluation (our goal is to evaluate {\em descriptors}), we use the ground-truth dense reconstruction to determine ground-truth correspondences (Sect. \ref{sect-dataset}) between training and test images, including scale. We then extract a patch of the same size as used in training from the same pyramid level of the test image. The test descriptors are then computed on the neighborhood of the true corresponding location. For each object, this generates around a thousand positive test descriptors, after rejecting those that are non-covisible. 

\subsection{Evaluation and Comparison}
\label{sect-eval}
Given a descriptor database, the simplest method to match a test query is via {\em nearest neighbor} (NN) search. While not sophisticated comparison method, this suffices to our purpose. We compare four combinations using the same NN search method: (1) SVHOG -- single view HOG computed on a random image from training sequence, (2) MV-HOG -- time-aggregated densely computed histogram of gradient orientation using all tracked patches, (3) KeepALL -- single view HOG descriptors computed and stored at each frame, and (4) R-HOG -- single view HG computed via the reconstruction, as described in Sect. \ref{sect-recon}.  The score used is {\em recognition rate}, shown in Fig. \ref{fig-vsall} for all 4 methods.

When averaging over all real objects and multiple random trials of the latter, MV-HOG improves accuracy by  $14\%$ ($16\%$ in synthetic objects) over SVHOG, and is only slightly worse than KeepALLHOG, but at a fraction of the cost. R-HOG nearly matches the performance of KeepALLHOG and improves on MV-HOG. Our implementation uses the range estimated by an RGB-D sensor, so it should be considered an upper-bound of performance, but it only uses $10$ keyframes per object, so as to keep the max-out cost in check. Run-time cost is reported in Sect. \ref{sect-cost}.

The fact MV-HOG and R-HOG perform similarly is indicative of the training sequences in the dataset being {\em sufficiently exciting,} as the given views are informationally equivalent to the reconstruction. This wold not be the case for short-baseline video, where the marginalized descriptor R-HOG can generate synthetic vantage points not represented in the training sequence, whereas MV-HOG would fail to capture a representative sample of the class-conditional distribution (Fig. \ref{fig-exc}). 

\begin{figure*}
\begin{center}
\includegraphics[height=0.3\textwidth,width=\textwidth]{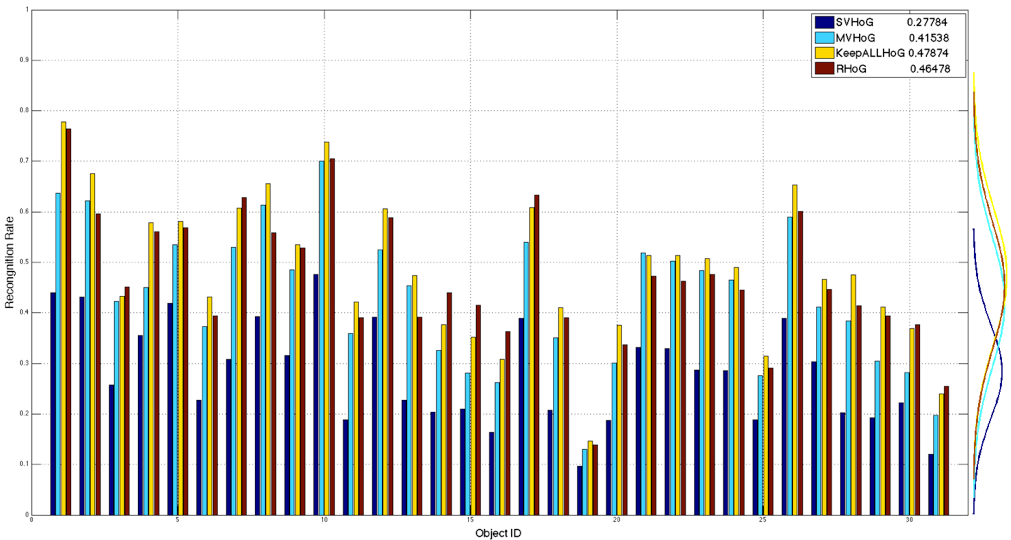}
\end{center}
\caption{ {\sl Comparison } Recognition rate for each of the 4 descriptors and each test object in the real dataset (abscissa), for a patch sizes $11\times 11$.The average accuracy score is shown in the legend. Distributions of recognition rate are shown on the right side. MV-HOG and R-HOG perform comparably to KeepAllHOG, but at a cost comparable to SV-HOG.}
\label{fig-vsall}
\end{figure*}

\begin{figure}
\begin{center}
\includegraphics[height=0.5\textwidth,width=\textwidth]{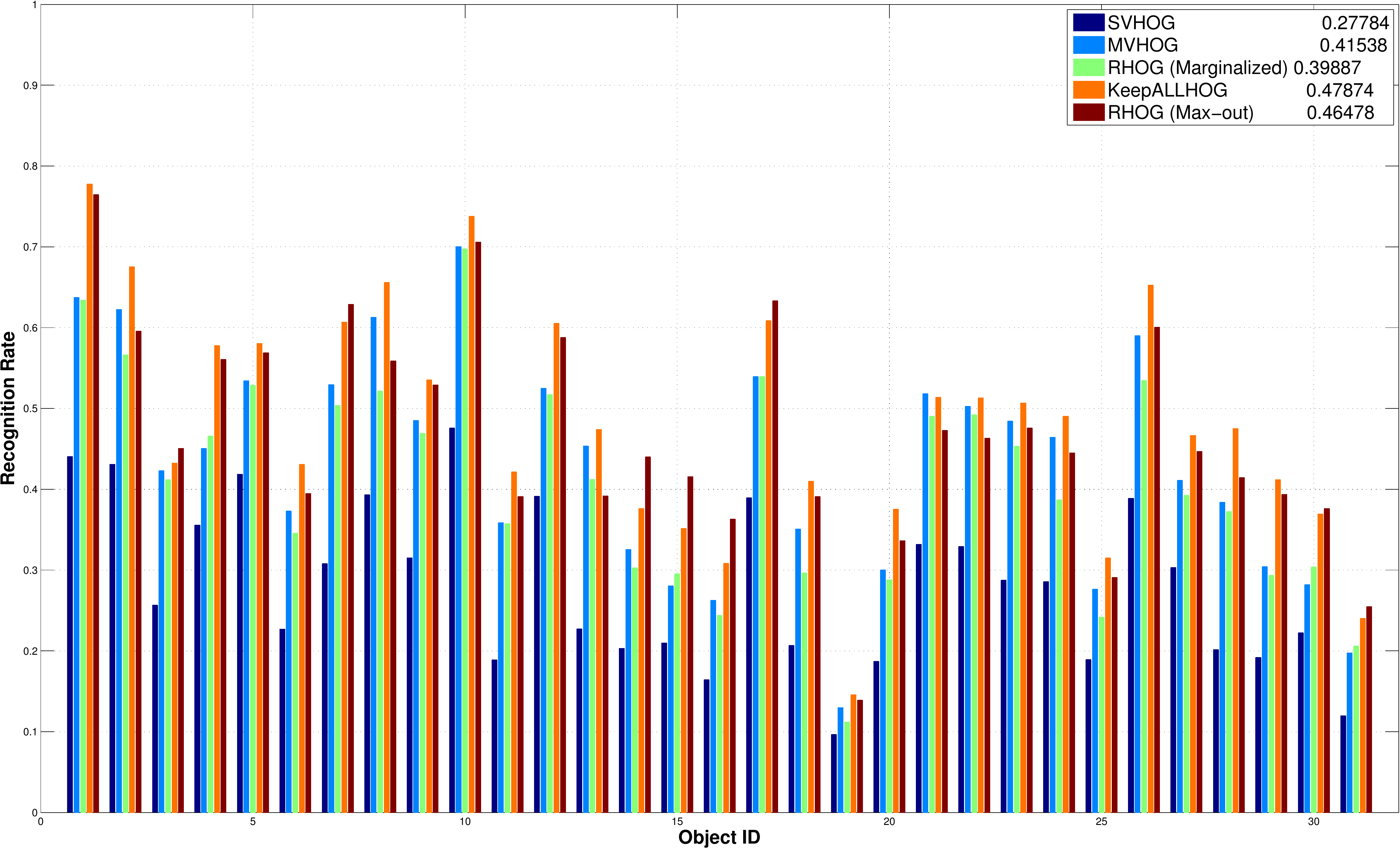}
\end{center}
\caption{ {\sl Comparison (Real, Patch Size $11 \times 11$)} MV-HOG outperforms SV-HOG in all cases and is slightly worse than the KeepALL strategy, but the latter induces significant computational cost at decision time. R-HOG is generated using a reconstruction computed from $10$ out of a total of $400$ frames in the training video. The performance is comparable toMV-HOG and KeepALL-HOG, but slightly inferior due to the inaccuracy of the reconstruction. Unsurprisingly, where the reconstruction improves, so does overall performance; this is visible in Fig. \ref{fig-syn_11x11} where reconstruction from synthetic data is more accurate.}
\label{fig-real_11x11}
\end{figure}

\begin{figure}
\begin{center}
\includegraphics[height=0.5\textwidth,width=\textwidth]{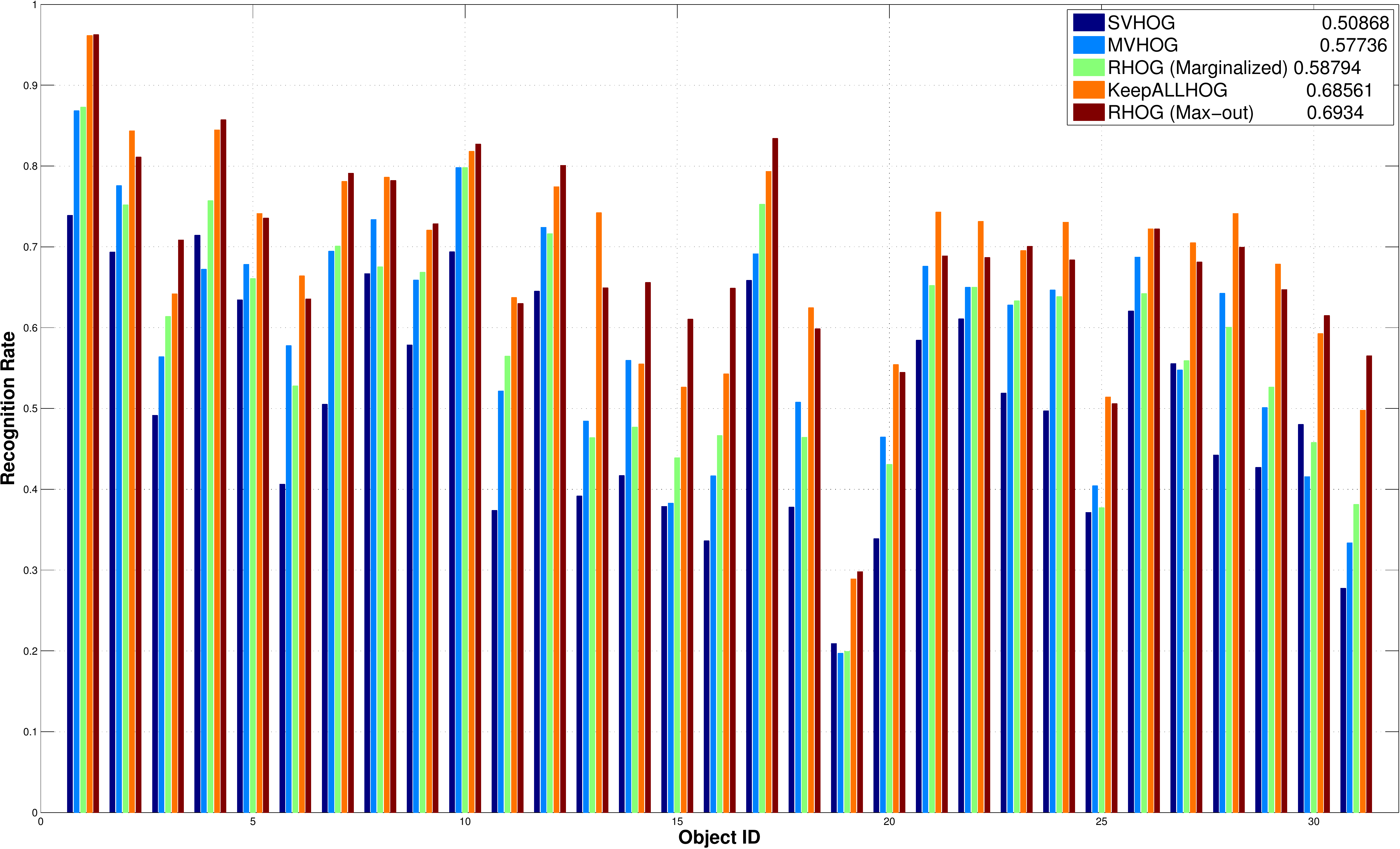}
\end{center}
\caption{ {\sl Performance of various descriptors on the Real dataset, as in Fig. \ref{fig-real_11x11} but for patch size $21 \times 21$. Although the numerical scores are different, the relative performance ranking are largely unchanged.}}
\label{fig-real_21x21}
\end{figure}

\begin{figure}
\begin{center}
\includegraphics[height=0.5\textwidth,width=\textwidth]{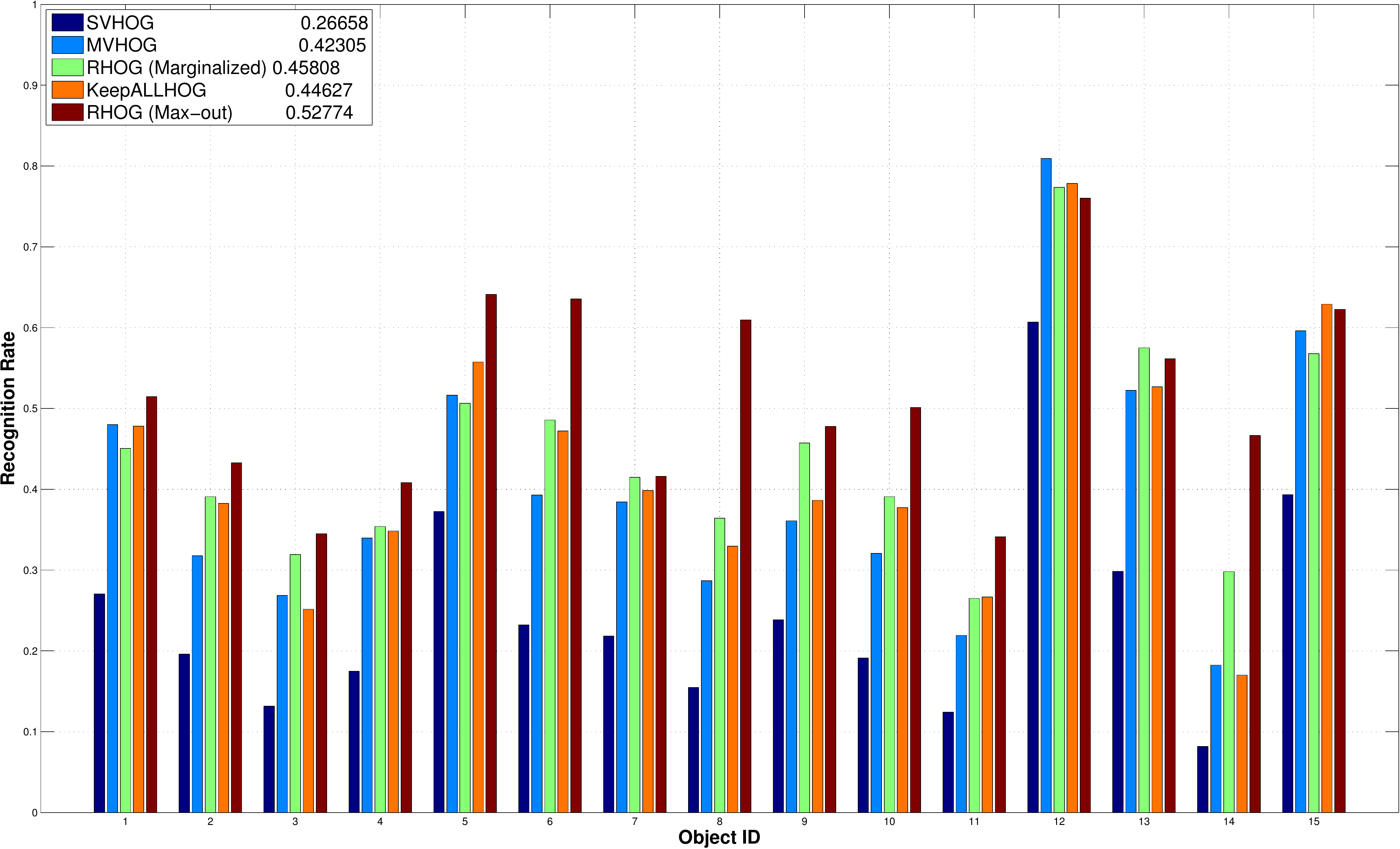}
\end{center}
\caption{ {\sl Comparison on the Synthetic dataset, for patch Size $11 \times 11$. Similar to the real dataset, MV-HOG outperforms SV-HOG in all cases and is similar to KeepALLHOG at a fraction of the computational cost. Both RHOG's are generated using $1$ out of a total $100$ frames in training video. Unlike Fig. \ref{fig-real_11x11}, reconstruction from synthetic data is more accurate, thus boosting performance in both marginalized and max-out descriptors.}}
\label{fig-syn_11x11}
\end{figure}

\begin{figure}
\begin{center}
\includegraphics[height=0.5\textwidth,width=\textwidth]{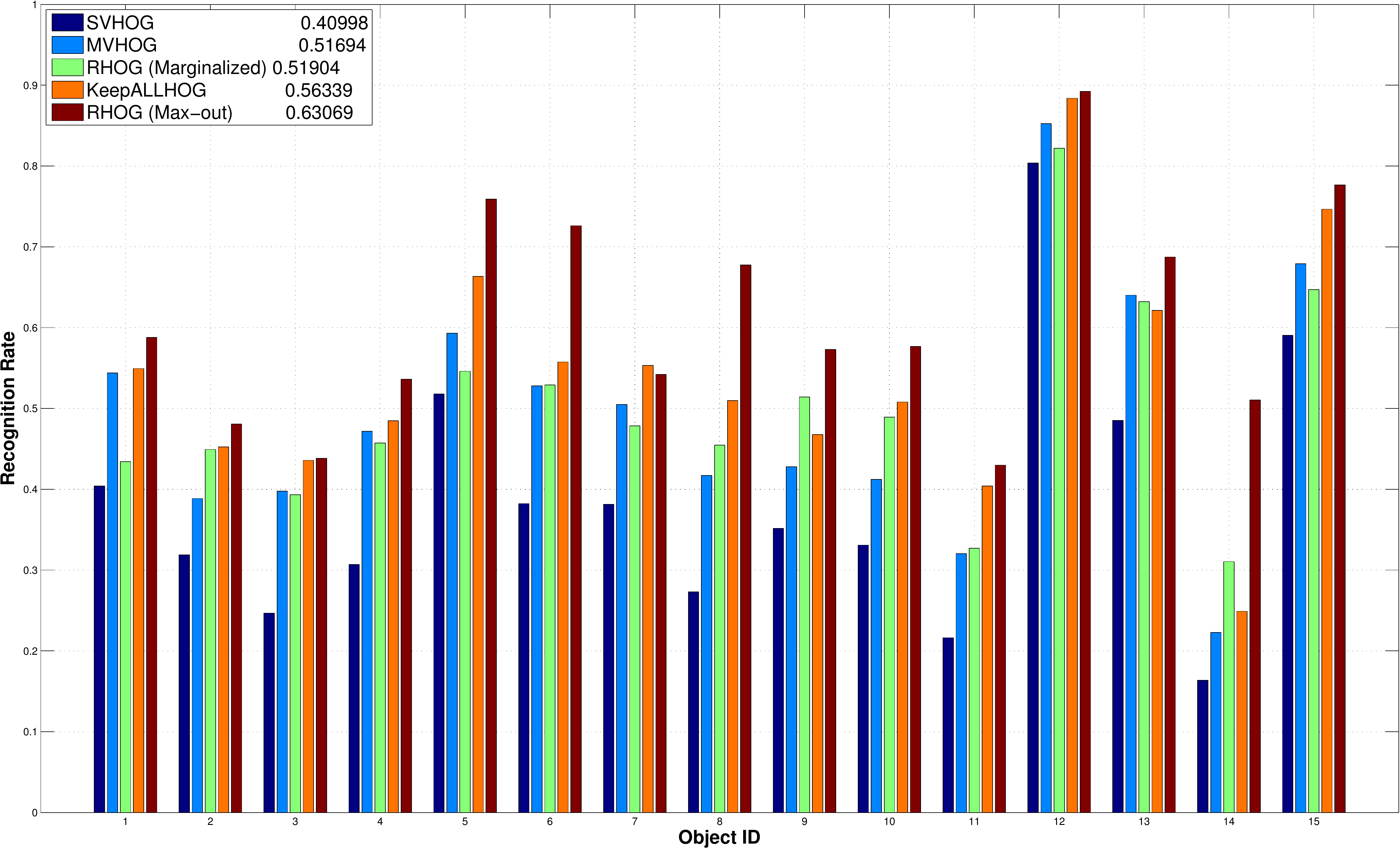}
\end{center}
\caption{ {\sl Comparison on the Synthetic dataset using patch Size $21 \times 21$. Performance is qualitatively similar to Fig. \ref{fig-syn_11x11}, although the numerical scores are different.}}
\label{fig-syn_21x21}
\end{figure}

Although the numerical scores change as we change the support regions (here we show the extrema of the range of the experiments we conducted, between $11\times 11$ and $21\times 21$ pixels), the conclusion holds across experiments: The performance of MV-HOG exceeds SV-HOG computed on a random image in the sequence, and is comparable to Keep-All HOG, despite requiring a fraction of the storage and computational cost. R-HOG performs comparably to MV-HOG so long as the reconstruction is accurate, and improves with the latter. This is visible in the comparison of the performance in the Real and Synthetic dataset. 

The numerical scores in the performance increase with the size of the patches, which is to be expected and may induce one to favor large patches, all the way to the entire image ({\em i.e.,} computing a single, dense HOG descriptor). This, however, is not viable as the conditions under which the descriptors are an approximation of the class-conditional densities involve the domain of the descriptor {\em not straddling an occlusion}. Since occlusions cannot be determined in a single image, one is left with guessing the size of the domain that trades off the probability of straddling an occlusion with the maximization of discriminative power. This issue can be addressed by computing the descriptors at multiple scales, or ``growing'' the size of the domain during the matching process; both approaches have been well explored in the literature and therefore are not further elaborated here.

\subsection{Time Complexity}
\label{sect-cost}
At test time, all descriptors have the same complexity. KeepALLHOG, on the other hand, needs every instance seen in the training sequence, so storage complexity grows from $O(n)$ to $O(kn)$ where in our case $k = 10$ and $n>10,000$ is the number of features stored. If evaluation of MV-HOG and R-HOG is done with max-out (Sec. \ref{sect-comparison}), the search approaches the complexity of KeepALLHOG. As a limit test, one can store the entire collection of (contrast-normalized) patches, and then search the entire dataset at test time. This can be done in approximate form using {\em approximate nearest neighbors} by preprocessing them into a search-efficient structure. Fig. \ref{fig-vsflann} shows the training time using the {\em fast library for approximate nearest neighbors} (FLANN) vs MV-HOG on a commodity PC with 8GB memory and Xeon E3-1200 processor. MV-HOG scales well and is more memory-efficient while KeepALLHOG (training using FLANN) requires more time and occupies more than 60\% of the available memory. Another advantage of MV-HOG is that the descriptor can be updated incrementally, and does not require storing processed samples for update. Fig. \ref{fig-vsall} shows that the performance loss of MV-HOG compared to KeepALLHOG is relatively small. 

\begin{figure}[t]
\begin{center}
\includegraphics[height=0.2\textwidth,width=0.9\linewidth]{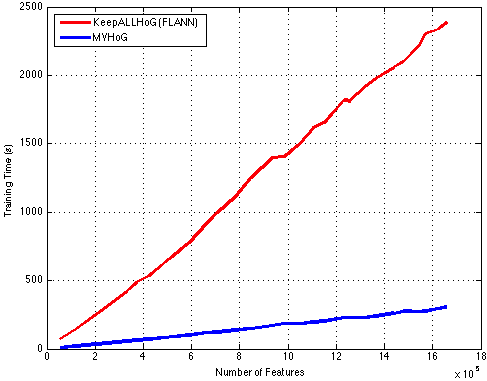}
\end{center}
\caption{{\sl Complexity Comparison.} Time complexity as a function of the number of features. FLANN precision is set to $0.7$. Higher precision will further increase computational load.}
\label{fig-vsflann}
\end{figure}

\subsection{Sample Sufficiency}

MV-HOG relies on a sufficiently exciting sample from the class-conditional being available in the training set. Clearly, if a sequence of identical patches is given (video with no motion), the descriptor will fail to capture the representative variability of images generated by the underlying scene. In this case, MV-HOG reduces to HOG. In Fig. \ref{fig-exc} we explore the relation between performance gain and excitation level of training sequence. As a proxy of the latter, we measure the variance of intensity to the mean patch using the $\ell_2$ distance. The right plot shows that the variance reaches the maximum when most frames are seen. We normalize the variance so that $1$ means maximum excitation. The left plot shows accuracy increases with excitation. The fact that accuracy does not saturate is due to the fact that the sufficient excitation is only reachable asymptotically. 

\begin{figure}[t]
\begin{center}
\includegraphics[width=0.9\linewidth]{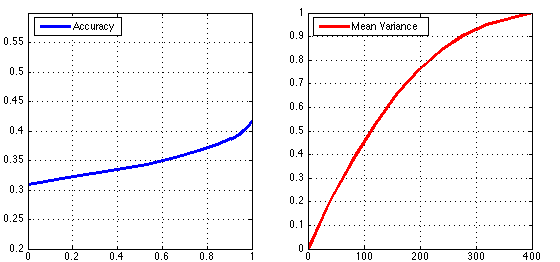}
\end{center}
\caption{{\sl Sufficient excitation.} Left: accuracy as a function of a proxy of sufficient excitation (see text). Right: Excitation as a function of the number of frames. All results are averaged over multiple runs using frames $i, \dots, i+k-1$ where $i$ is selected at random.}
\label{fig-exc}
\end{figure}

\section{Discussion}
\label{sect-discussion}

By interpreting HOG as the probability density of sample images, conditioned on the underlying scene, with nuisances marginalized, and observing that a single image does not afford proper marginalization, we have been able to extend it using nuisance distributions learned from multiple training samples. The result is a multi-view extension of HOG that has the same memory and run-time complexity of its single-view counterpart, but better trades off sensitivity to discriminative power, as shown empirically.

Our method has several limitations: It is restricted to static (or slowly-deforming) objects; it requires correspondence in multiple views to be assembled (although it reduces to standard HOG if only one image is available), and is therefore sensitive to the performance of the tracking (MV-HOG) or reconstruction (R-HOG) algorithm. The former also requires sufficient excitation conditions to be satisfied, and the latter requires sufficiently informative data for multi-view stereo to operate, although if this is not the case, then by definition the resulting descriptor is insensitive to nuisance factors; it is also, of course, uninformative, and therefore this case is of no major concern. It also requires the camera to be calibrated, but for the same reason, this is irrelevant as what matters is not that the reconstruction be correct in the Euclidean sense, but that it yields consistent reprojections. 

Our empirical evaluation of R-HOG yields a performance upper bound, as we use the reconstruction form a structured light sensor rather than multi-view stereo. As the quality (and speed) of the latter improve, the difference between the two will shrink.

\end{document}